From Detection to Discovery: A Closed-Loop Approach for Simultaneous and Continuous Medical Knowledge Expansion and Depression Detection on Social Media

Shuang Geng, Ph.D.
Shenzhen University
1066 Xueyuan Road
Shenzhen, China 518055
Email: gs@szu.edu.cn

Wenli Zhang, Ph.D.
Iowa State University
3332 Gerdin, 2167 Union Drive
Ames, IA, USA 50011-2027
Email: wlzhang@iastate.edu

Jiaheng Xie, Ph.D.
University of Delaware
217 Purnell Hall
Newark, DE, USA 19716
Email: jxie@udel.edu

Rui Wang
Shenzhen University
1066 Xueyuan Road
Shenzhen, China 518055
Email: wangrui2023@email.szu.edu.cn

Sudha Ram
University of Arizona
McClelland Hall 430J, 1130 E. Helen St.
Tucson, Arizona 85721-0108
Email: ram@eller.arizona.edu

Please send comments to Wenli Zhang at wlzhang@iastate.edu.

# From Detection to Discovery: A Closed-Loop Approach for Simultaneous and Continuous Medical Knowledge Expansion and Depression Detection on Social Media

**ABSTRACT**

Social media user-generated content (UGC) provides real-time, self-reported indicators of mental health conditions such as depression, offering a valuable source for predictive analytics. While prior studies integrate medical knowledge to improve prediction accuracy, they overlook the opportunity to simultaneously expand such knowledge through predictive processes. We develop a Closed-Loop Large Language Model (LLM)–Knowledge Graph framework integrates prediction and knowledge expansion in an iterative learning cycle. In the knowledge-aware depression detection phase, the LLM jointly performs depression detection and entity extraction, while the knowledge graph represents and weights these entities to refine prediction performance. In the knowledge refinement and expansion phase, new entities, relationships, and entity types extracted by the LLM are incorporated into the knowledge graph under expert supervision, enabling continual knowledge evolution. Using large-scale UGC, the framework enhances both predictive accuracy and medical understanding. Expert evaluations confirmed the discovery of clinically meaningful symptoms, comorbidities, and social triggers complementary to existing literature. We conceptualize and operationalize prediction-through-learning and learning-through-prediction as mutually reinforcing processes, advancing both methodological and theoretical understanding in predictive analytics. The framework demonstrates the co-evolution of computational models and domain knowledge, offering a foundation for adaptive, data-driven knowledge systems applicable to other dynamic risk monitoring contexts.

*Keywords*: closed-loop learning, knowledge-driven machine learning, knowledge expansion, healthcare analytics, user-generated content

# 1 INTRODUCTION

User-generated content (UGC) on social media contains an abundance of real-time, self-reported signals about individuals' psychological states, behaviors, and life events (Chau et al., 2020; W. Zhang et al., 2024). This rich, naturally occurring data has fueled a growing body of research on detecting depression on social media using machine learning and natural language processing (NLP), so that platforms can develop new services with personalized recommendations and create supportive environments for those affected by depression (Chau et al., 2020; Ghahramani et al., 2022; W. Liu et al., 2020; Yang et al., 2022; D. Zhang et al., 2024; W. Zhang et al., 2024).

A common strategy in these efforts is to incorporate external domain knowledge, such as medical ontologies or lexicons, into predictive models to enhance accuracy (Chau et al., 2020; D. Zhang et al., 2024; W. Zhang et al., 2024). Depression detection from UGC particularly benefits from these medical knowledge-aware approaches because purely data-driven methods often overfit to dataset-specific lexical cues and fail under domain, demographic, or temporal shifts (Harrigian & Dredze, 2022). These challenges are amplified when depression signals in UGC are sparse, heterogeneous, and highly context-dependent (e.g., comorbidity, life events, culture) (W. Zhang et al., 2024). By embedding domain knowledge, knowledge-aware methods constrain learning, enhance robustness to drift and label noise, and enable more data-efficient prediction. While effective, these approaches remain fundamentally *open-loop*: they consume medical knowledge but do not contribute new insights back to the knowledge base. As a result, the vast potential of UGC to enrich clinical understanding of depression (by surfacing emerging symptoms, novel comorbidities, and evolving social triggers) remains underutilized.

We argue that progress in this area requires moving beyond prediction as an endpoint and adopting a *closed-loop* framework that organically integrates prediction and knowledge expansion into a continuous cycle (Figure 1). In a closed-loop system, prediction and knowledge expansion are mutually reinforcing rather than separate tasks. Existing medical knowledge improves prediction, while prediction, applied to large-scale UGC, generates new knowledge that can be validated and fed back into the knowledge base. This feedback loop not only improves the performance of depression detection models over successive iterations but also contributes to advancing domain knowledge itself. This study operationalizes this vision by developing a Closed-Loop Large Language Model (LLM)–Knowledge Graph Framework that jointly learns depression detection and medical knowledge expansion (i.e., new entities, entity relationships, and entity types) from UGC.

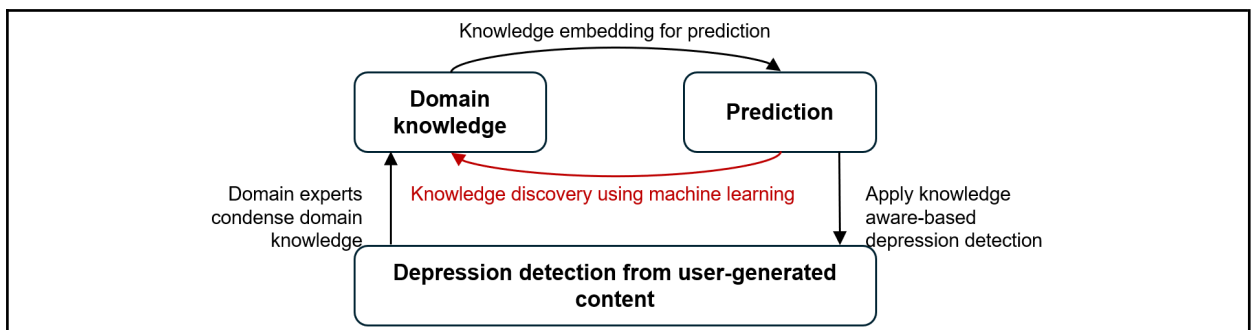

**Figure 1. From Detection to Discovery: Predicting Through Learning and Learning Through Predicting**

A critical question is why prediction and knowledge discovery need to be pursued together within the same framework rather than separately as independent and unconnected tasks. Treating these tasks in isolation risks missing their mutually reinforcing nature. Prediction without accompanying knowledge expansion relies on static domain knowledge and lacks the capacity to adapt to emerging patterns in UGC (e.g., self-reported new types of symptoms, evolving comorbidities, or real-world events as new triggers of depression across populations) (Sermo & LiveWorld, 2023). Conversely, knowledge expansion without integration into prediction tasks risks producing fragmented insights that remain disconnected from practical applications, without evidence that the new knowledge can actually improve machine learning performance. Integrating prediction and knowledge expansion within the same closed-loop framework allows us to demonstrate how new knowledge both arises from and enhances predictive performance. This integrated design makes it possible to evaluate the loop as a system, showing that its value lies not in prediction or knowledge expansion alone, but in the continuous feedback cycle between the two.

Designing such a framework, however, poses significant challenges. The first challenge lies in the nature of depression-related domain knowledge. Symptoms and risk factors often appear in fragmented forms across different scales, in the literature, and in descriptive accounts. However, in prediction tasks, it is essential to recognize that depression symptoms and factors are not isolated. They are interconnected and may influence one another in complex ways, including both interactions among entities and interactions among the relationships between entities (for example, a medication (e.g., fluoxetine) may be taken during a life event (e.g., dealing with a crisis), while a psychological symptom (e.g., anxiety) is triggered by that event and co-occurs with another symptom (e.g., aboulomania)). Modeling these intricate connections and incorporating them effectively into depression detection remains a substantial challenge. The second challenge concerns the discovery of new knowledge from UGC. While it is possible to identify novel depression-related risk factors, symptoms, and other elements in the form of entities or relations, it is often unclear how these new findings relate to existing knowledge, whether they are valid, and how they can be effectively leveraged in applications. One potential strategy is to leverage labeled data from the prediction process to guide the discovery of new knowledge in UGC. However, operationalizing this approach has not yet been explored. The third challenge is how to integrate knowledge-aware prediction and knowledge discovery into a unified framework where both processes can reinforce and validate each other.

This study addresses these challenges by using depression as a research case and pursuing three interrelated goals. First, we aim to develop a *continuous loop* that integrates knowledge-aware depression detection with the expansion of medical knowledge on depression. Second, we seek to design a pathway that quantifies the importance of depression-related entities for prediction from a knowledge graph, and leverages domain knowledge to improve model performance. Third, we aim to establish a knowledge refinement and expansion pathway that uses data from both depressed and non-depressed users utilized during depression detection to refine the knowledge graph for practical utility, to discover new medical knowledge (depression-related entities, their relationships, and new entity types), to incorporate expert validation, and to seamlessly reintegrate validated insights into subsequent prediction cycles.

Guided by computational design science principles (Gregor & Hevner, 2013; Hevner et al., 2004), we develop a Closed-Loop LLM–Knowledge Graph Framework to operationalize this vision. In our approach, we use a LLM that can understand and analyze language to learn two tasks simultaneously: identifying depression-related entities and detecting depression from UGC. We fine-tune the LLM using Low-Rank Adaptation (LoRA), an efficient method that enables the model to learn new tasks without being retrained from scratch. As the model learns, it discovers new depression-related entities that are added to a medical knowledge graph, a structured representation showing how depression-related factors are connected. The attention network within the knowledge graph then evaluates the importance of each new piece of information for improving depression detection. This updated knowledge is incorporated into the next training iteration, forming a continuous learning loop that enhances both the predictive model (its ability to detect depression) and the knowledge graph (its medical understanding of depression). Specifically, our objective function jointly optimizes the losses for depression-related entity recognition, depression detection, and knowledge refinement.

Positioned in the machine learning research in information systems (IS) (Padmanabhan et al., 2022), our work represents a Type I contribution focused on method development by introducing a new framework for iterative loops of prediction and knowledge discovery. More broadly, our design highlights a new problem space in IS research: how to continuously integrate and expand domain knowledge in tandem with machine learning models. This work provides salient design insights for future research while addressing the societal challenge of underdiagnosed depression by offering a tailored machine learning framework that both detects depression effectively and advances medical knowledge. Practically, our proposed framework can assist social media platforms and related organizations in identifying users at risk of depression and consequently provide relevant online support information. Additionally, because our framework can uncover new knowledge related to depression from UGC on social media, it can supplement existing knowledge that is difficult to obtain in traditional scientific or clinical studies. This, in turn, helps medical professionals gain a more comprehensive understanding of depression-related factors and their evolution within populations, which may also pave the way for new scientific directions.

## 2 RELATED WORK

### 2.1 Prediction and Knowledge Expansion: Open-Loop vs. Closed-Loop Approaches

As mentioned in the Introduction, integrating domain knowledge into depression detection models in UGC can enhance predictive performance. At the same time, UGC provides opportunities to uncover new depression-related knowledge. Research at this intersection has produced two largely separate streams: *prediction-oriented studies* and *knowledge discovery–oriented studies*. These approaches lack integration, as there is no existing framework that effectively links prediction with knowledge expansion. Treated independently, prediction depends on static knowledge and cannot adapt to emerging new knowledge, while knowledge expansion risks generating fragmented insights without clear utility. The central technical challenge lies in developing an integrated approach that *closes the loop.* This integration is non-trivial: the prediction process can inform the identification of new knowledge, and in turn, newly identified knowledge can be incorporated into prediction models, validated through improvements in predictive performance, and reintegrated into subsequent cycles.

For prediction-oriented studies, researchers often leverage existing medical ontologies and knowledge graphs to enhance diagnostic or prognostic models. For instance, Jiang et al. (2023) construct patient-specific knowledge graphs from electronic health records to enhance mortality prediction, while Yang et al. (2022) and Tavchioski et al. (2022) enrich social media text with commonsense or latent knowledge graph features to improve depression prediction. These works show that injecting structured knowledge can significantly boost model performance. Yet they treat the knowledge source as *static*, assuming that medical concepts and relationships remain fixed, which limits the model's ability to adapt to emerging patterns in UGC. By contrast, knowledge expansion studies emphasize the construction or expansion of medical knowledge. Gao et al. (2025) introduce the Mental Disorders Knowledge Graph, which integrates biomedical literature into a graph of over 10 million relations, including nearly 1 million novel associations, and demonstrates that its use accelerates expert validation by up to 70%. Manzoor et al. (2023) deploy a human-in-the-loop framework with UGC, showing that their approach expanded the knowledge graph four times faster than manual methods and contributed to a 20% increase in ad revenue. These efforts underscore the dynamic and evolving nature of knowledge expansion, but they rarely demonstrate how newly generated knowledge can be dynamically utilized in predictive tasks.

Our proposed Closed-Loop LLM–Knowledge Graph framework diverges from prior studies by integrating prediction and knowledge expansion into a *continuous, iterative cycle*. Unlike static prediction models (Jiang et al., 2023; Yang et al., 2022) or standalone knowledge expansion efforts (Gao et al., 2025), our approach demonstrates how new knowledge both arises from and enhances predictive performance. This shift moves beyond incremental improvements in accuracy or knowledge coverage: it reframes UGC-based mental health analytics as a dynamic system in which prediction and discovery are inseparable, reinforcing, and co-evolving. Such a design not only improves predictive accuracy (through the dynamic updating and integration of knowledge graph information into subsequent predictions) but also enriches actionable medical knowledge (through the refinement and expansion of the knowledge graph itself).

Several existing studies have also emphasized the importance of closed or continuous loops of knowledge discovery, but they differ fundamentally from our approach. Girardi et al. (2016) put the *doctor-in-the-loop* by redesigning the knowledge discovery workflow for biomedical research so clinicians iteratively guide data modeling, validation, analysis, and ontology updates; crucially, the process includes explicit feedback loops (e.g., analysis → data and ontology revision) demonstrated on cerebral-aneurysm data: an expert-centered, interactive loop rather than a purely automated one. Musslick et al. (2025) argue for *closed-loop scientific discovery* in behavioral science that algorithmically integrates hypothesis generation, experimental design, data collection, model updating, and next-experiment selection into a continuous cycle, aiming to automate the science loop itself. Saikin et al. (2019), in drug discovery, call for *platform-level integration* that links generative/inverse-design models with automated synthesis and high-throughput assays, so measurements feed back to models, i.e., a lab-to-model-to-lab loop needed for AI to have a practical impact. Rather than a conceptual/process call (Saikin et al., 2019) or expert-interactive loop (Girardi et al., 2016) or a field-level automation vision (Musslick et al., 2025), our framework *operationalizes* a healthcare-specific approach of predicting through learning and learning through predicting using UGC: it (*i*) mines population-scale UGC to expand a medical knowledge graph, (*ii*) *immediately* reintegrates the vetted additions via knowledge embeddings into depression detection, and (*iii*) uses realized predictive gains as system-level evidence to govern subsequent knowledge updates, i.e., closing the loop between knowledge expansion and prediction in the same deployable pipeline.

**2.2 UGC for Depression Detection and Medical Knowledge Expansion**

Extensive efforts have been devoted to understanding depression, particularly in acquiring knowledge about its symptoms and risk factors. In the medical domain, there are two distinct approaches to knowledge acquisition, such as identifying depression symptoms and risk factors. One way is through scientific or clinical studies, which often provide information crucial for detecting and treating depression (Adams et al., 2025). Another is to conduct long-term, continuous, large-scale observations of population behaviors and mental traits to identify correlations with depression and ultimately synthesize these findings into medical knowledge (Beck & Alford, 2014). For a long time, the latter method has been difficult to implement due to constraints in time, manpower, and financial resources, as such studies typically span over years or even decades. With the abundance of UGC, such observational studies could potentially be carried out in a more cost-effective and accessible manner: Kallinikos & Tempini (2014)'s work highlights that UGC can break the strong expert culture of medical studies, allowing for the capture of patient life details that have traditionally remained outside the medical research's scope and can serve as a complement to established medical research methods. For example, social media is influencing clinical judgment (Jilka & Giacco, 2024). 57% of U.S. physicians report that content they see on social platforms occasionally changes their perceptions of medications and treatments (Sermo & LiveWorld, 2023).

Existing approaches, however, use UGC for understanding depression (or mental disorders in general) in different and largely disconnected ways. One line of work treats UGC primarily as input for identifying individuals with depression. Within this line of work, a key direction focuses on enriching social media text with external domain knowledge to improve depression classification (Table 1). This stream of research employs four types of knowledge integration strategies. First, rule or lexicon-guided hybrids combine machine learning with expert rules or domain lexicons, enhancing precision on high-risk cues (Chau et al., 2020). Second, knowledge-aware sequence models infuse external commonsense or mental-state knowledge into deep learning models to capture latent reasoning patterns (Yang et al., 2022). Third, ontology and knowledge graph-driven pipelines extract diagnosis-related entities, organize them into medical ontologies, and use knowledge-aware detectors for classification (W. Zhang et al., 2024). Fourth, language model knowledge enhancement incorporates knowledge graphs into model representations, dynamically aligning embeddings or injecting structured triples into pretrained LLMs (W. Liu et al., 2020; Zhong et al., 2019).

The trajectory above shows steady gains from hand-crafted rules, to knowledge-aware deep models, and to language models with knowledge enhancement. The emergence of contemporary LLMs has propelled this methodological shift by offering stronger language understanding, entity and relation extraction, and flexible reasoning that can both use and expand domain knowledge (e.g., extracting candidate entities and relations to update a knowledge graph or ontology, as in our closed-loop design). These advances make LLMs practical to tailor for mental health-related healthcare analytics at both research and deployment scales, motivating a closer review of LLMs for developing our closed-loop framework.

| Table 1. Representative Work on Using UGC with Domain Knowledge for Depression Detection | | | | |
|---|---|---|---|---|
| References | Task/domain | Knowledge source/type & Knowledge's role | How knowledge is integrated | Learning method |
| (Chau et al., 2020) | Identifying emotionally distressed bloggers (Chinese) | - Expert-authored rules (domain knowledge)<br>- Rules inject domain priors and improve utility/precision in high-risk retrieval | Hybrid design: combine machine learning classification with rule-based classification derived from experts | Traditional machine learning and system-level hybrid |
| (Yang et al., 2022) | Stress & depression detection from social posts | - Commonsense/mental-state knowledge from COMmonsEnse TransFormer<br>- External commonsense makes latent mental-state reasoning more explicit for classification | Infuse knowledge via GRUs; knowledge-aware mentalisation (dot-product attention) to attend to relevant knowledge; supervised contrastive learning for class-specific features | Knowledge-aware RNN and attention with contrastive objective |
| (W. Zhang et al., 2024) | Depression detection from social-media traces | - Depression diagnosis-related entities (symptoms, life events, treatments) and depression ontology<br>- Medical knowledge is operationalized as entities and an ontology that guide the final detector | Three modules: (1) NER extracts diagnosis-related entities (RoBERTa-based, transition/Stack-LSTM architecture); (2) construct/domain ontology; (3) knowledge-aware attention-based sequence model for detection | Pipeline and knowledge-aware detector (deep learning) |
| (D. Zhang et al., 2024) | Suicidal ideation detection from social media; generalization tests | - Suicidal ideation lexicon (domain terms)<br>- Lexicon steers the Transformer's representation toward domain-salient cues; validated across languages/platforms | Model-level integration of a social-media suicidal ideation lexicon into a Transformer with aligned dynamic embeddings and lexicon-based enhancement capturing domain relevance + contextual importance | Knowledge-enhanced Transformer |
| (Zhong et al., 2019) | Emotion detection in conversations (social media settings) | - External commonsense (e.g., ConceptNet/affective knowledge)<br>- Shows consistent gains from injecting commonsense for social text semantics | Context-aware affective graph attention to incorporate external knowledge into Transformer context modeling | Knowledge-enriched Transformer |
| (W. Liu et | General | - Knowledge graph triples (e.g., domain | Inject knowledge graph triples as additional | Knowledge-en |
| al., 2020) | language understanding; widely used in social tasks | knowledge graphs)<br>- Provides a general recipe for structured-knowledge injection into PLMs used in social-text analytics | tokens with soft-position & visibility controls to fuse factual knowledge into BERT | abled BERT |

**Note**: UGC: User-generated content

Another line of using UGC for understanding depression (or other diseases) is to complement medical ontologies by extracting novel associations or patient-centered perspectives that extend formal knowledge bases (Table 2). Within this stream, three methodological directions can be distinguished: (*i*) lexicon building, where colloquial expressions are expanded and normalized to clinical ontologies (e.g., Reddit-trained drug lexicons and long-COVID symptom lexicons) (Lavertu & Altman, 2019), (*ii*) ontology mapping and relation extraction, such as identifying drug condition pairs (e.g., adverse reaction and symptom) and aligning them with resources like FDA adverse event reporting system (Smith et al., 2018), and (*iii*) UGC-derived knowledge graphs, which structure UGC (e.g., stressors, expectations, responses, affect) to support reasoning and downstream applications (Gao et al., 2025).

These approaches demonstrate how UGC, including patient language, symptoms, and relations, can be systematically structured to extend medical knowledge. We also observe a methodological shift toward knowledge graph-based knowledge representations. This trend arises because UGC-derived medical knowledge is inherently relational and heterogeneous, and knowledge graphs are particularly effective in this context: they capture multi-hop dependencies while re-weighting edges by importance. This suitability motivates our subsequent review of knowledge graph attention networks.

| Table 2. Representative Work on Using User-Generated Content to Extend Medical Knowledge | | | |
|---|---|---|---|
| **References** | **UGC source** | **Knowledge representation** | **Methods & how it complements medical knowledge** |
| (Kallinikos & Tempini, 2014) | Patient self-reporting networks | Community-curated patient data | IS qualitative analysis; shows UGC as "medical facts" alongside expert data |
| (Lavertu & Altman, 2019) | Reddit (health subreddits) | Lexicon and embeddings | Word embeddings; expands drug vocabulary beyond ontologies |
| (Nikfarjam et al., 2015) | Health social media | ADR mentions | Conditional random fields and embeddings; extracts drug–ADR relations for pharmacovigilance |
| (Smith et al., 2018) | Twitter (adalimumab posts) | ADRs mapped to UMLS | Semi-automatic extraction and UMLS mapping; complements FAERS |
| (Sarker & Ge, 2021) | Reddit (/r/covidlonghaulers) | Symptom lexicon to ontology IDs | Lexicon matching; surfaces long-COVID symptom knowledge |
| (Welivita & Pu, 2022) | Reddit distress dialogues | UGC-derived KG (22k nodes, 104k edges) | Topic modeling and clustering; builds knowledge graph of stressors/affect |

| (Gao et al., 2025) | Literature + curated DBs | Multi-relational KG (MDKG) | LLM-assisted knowledge graph construction; expands mental-health knowledge |

**Note**: UGC: User-generated content; ADR: Adverse Drug Reaction; UMLS: Unified Medical Language System; KG: Knowledge graph

To summarize, our review of the literature shows that prediction-oriented studies often overlook the knowledge-discovery potential of UGC, whereas knowledge-extraction efforts frequently stop at identifying candidate new knowledge without demonstrating their downstream benefits for prediction or clinical decision-making. Our objective differs from prior UGC-based work: we propose a closed-loop machine learning framework that continuously and automatically enriches the medical knowledge base with insights derived from UGC, then reintegrates and validates this expanded knowledge within predictive tasks for depression detection. In doing so, prediction and knowledge discovery are pursued together, allowing newly discovered knowledge to demonstrably enhance predictive performance and, in turn, guiding further discovery. Realizing this vision raises non-trivial challenges. UGC is vast, heterogeneous, and noisy, with much of it unrelated to mental health; isolating medically meaningful signals requires robust filtering and modeling. Moreover, aligning emergent insights with established medical ontologies, while ensuring clinical actionability, remains difficult. Addressing these challenges is central to leveraging UGC not only as a supplement to detection, but as a driver of medical knowledge expansion within a truly closed loop.

## 2.3 Related Machine Learning Methods

**Large Language Models and Low-Rank Adaptation Technique.** Prior literature demonstrates that LLMs are increasingly effective for mental health analytics, such as detecting depression and stress from UGC. Yet, when adaptation to new tasks beyond the capacity of general-purpose LLMs is required, it is neither feasible nor practical to train models from scratch or to perform large-scale full fine-tuning for every healthcare application (Hu et al., 2021). The barriers include enormous computational costs, substantial data requirements, and the difficulty of continuously updating models as new signals emerge (Hu et al., 2021). In our research context, i.e., closed-loop depression detection and knowledge expansion using UGC, LLMs must also address multiple tasks simultaneously, including prediction, depression-related entity recognition, and entity–relation extraction for knowledge expansion. These challenges collectively motivate our focus on low-rank adaptation (LoRA), which offers a parameter-efficient and modular strategy for tailoring LLMs to specialized healthcare tasks (Hu et al., 2021).

LoRA adapts LLMs by training a small set of additional parameters while freezing the base model. LoRA injects low-rank update matrices into Transformer layers, cutting trainable parameters by orders of magnitude with negligible inference overhead and performance comparable to full fine-tuning. In the clinical domain, Clinical LLaMA-LoRA adopts a two-stage Parameter-Efficient Fine-Tuning (PEFT) pipeline (Gema et al., 2024): one adapter specializes the LLM to clinical notes, and a second adapter targets downstream tasks; fusing both improves performance across multiple clinical NLP tasks while reducing computation, illustrating how adapter composition can support multi-objective pipelines. In biomedical relation extraction, which is directly relevant to our depression-entity discovery task, recent studies have reported the successful use of a LoRA variant (QLoRA) for low-resource relation extraction in underexplored biomedical domains (Delmas et al., 2024). This provides evidence that LoRA can effectively support knowledge extraction components even when only limited labeled data are available. Beyond healthcare, LoRA has been extended to multi-task settings (e.g., Mixture-of-LoRAs) to share representations while keeping task-specific deltas separate: patterns we leverage by tying entity identification and depression detection together with shared, low-rank modules (Feng et al., 2024). Collectively, these strands show that (*i*) LoRA enables frequent, low-cost updates; (*ii*) LoRA can compartmentalize knowledge and tasks; and (*iii*) LoRA works in medical knowledge extraction and classification settings.

Prior LoRA work in healthcare typically optimizes either knowledge extraction (e.g., relation extraction or entity recognition) or downstream prediction, and when both are present, treats them as loosely coupled stages (e.g., domain adapter and task adapter). In contrast, our design jointly learns depression-related entity recognition and depression detection within one LLM and LoRA framework and feeds the newly discovered entities back into a medical knowledge graph, where the knowledge graph attention network quantifies each entity's predictive utility; the updated knowledge is then re-injected into the next training cycle. LoRA's efficiency and modularity make this end-to-end closed loop feasible at the scale of large and continuously expanding UGC datasets. This design enables the integration of knowledge expansion and depression prediction into a single, iteratively executed system, rather than treating them as two independent tasks.

With the development of LLMs and agentic AI, one alternative approach is a multi-agent configuration in which a "predictor" agent and a "knowledge-discovery" agent iterate and exchange messages, optionally with human validation (Wu et al., 2023). However, this design adds orchestration overhead and, more importantly, does not address two persistent difficulties: *non-stationarity* (as agents keep changing each other's inputs, the learning target shifts) and *credit assignment* (when performance improves, it is hard to credit a *specific* knowledge edit). Recent surveys emphasize that these remain open challenges in multi-agent learning (Ning & Xie, 2024). These issues show up clearly in benchmark agent cooperative tasks (e.g., StarCraft battles where each agent controls one unit): Proximal Policy Optimization (PPO) variants are often adopted because their on-policy updates can help stabilize training when teammates' behaviors change, and value-decomposition methods explicitly target credit assignment by factorizing the joint action-value; yet both methods have known limitations, including confounding in decomposition and generalization gaps on difficult benchmarks tasks (W. Chen et al., 2023). Moreover, LLM-based multi-agent frameworks introduce additional failure modes in coordination and evaluation (Cemri et al., 2025).

These challenges motivate our design choices. Our closed-loop framework integrates knowledge refinement directly into a differentiable objective: entity recognition loss, prediction loss, and knowledge-refinement loss are jointly optimized, so knowledge graph updates receive immediate gradient-based supervision from the prediction task. In multi-agent systems using indirect reinforcement signals, the indirect reward signals often come after many rounds and are prone to high variance and delayed attribution, which slows convergence (Pignatelli et al., 2024). By contrast, direct supervision (loss or gradient) provides immediate, low-variance feedback that cleanly links a knowledge update to prediction improvement (X. Wang, Gao, et al., 2021). The framework enables stable updates and generates knowledge that is clearly useful for prediction, while also allowing domain experts to validate uncertain cases.

**Knowledge Graphs.** Knowledge graphs represent domain knowledge as networks of entities (nodes) and relations (edges), in a form that is both human- and machine-readable (Ji et al., 2022). In practice, curated knowledge graphs are often static snapshots: they encode what is known at build time and change only when experts or pipelines update them. This static nature limits their usefulness in settings where signals evolve or where task objectives require selectively emphasizing different parts of the graph.

Knowledge graph attention networks address this limitation by learning to weight nodes and relations dynamically for a given objective. During training, attention mechanisms prioritize the most informative neighbors and relation types, enabling (*i*) task-specific focus (e.g., emphasizing depressive-symptom neighbors for depression detection), (*ii*) adaptiveness via updated embeddings for entities and relations grounded in data, and (*iii*) end-to-end optimization for downstream prediction (Ji et al., 2022). Building on this idea, attention has diversified: hierarchical attention distinguishes importance at both node and semantic or meta-path levels in heterogeneous graphs (X. Wang, Ji, et al., 2021); relation-aware attention explicitly models relation types during neighborhood aggregation for knowledge graph completion (Z. Zhang et al., 2020a); transformer-style hierarchical attention first composes local entity-relation pairs and then aggregates across relations (S. Chen et al., 2021); and multi-hop attention propagates importance scores to capture dependencies beyond immediate neighbors (G. Wang, Ying, et al., 2021). These variants share the goal of aligning message passing with task-relevant structure.

The applications of knowledge graphs broadly span four areas (Ji et al., 2022) (Table 3): knowledge acquisition (extracting new entities, relations, and facts); representation learning (embedding entities and relations for reasoning); temporal or dynamic knowledge graphs (modeling time-varying states and links); and knowledge-aware applications (injecting knowledge graph signals into tasks such as recommendation, question answering, and healthcare prediction). Attention models cut across these areas by making inferences and learning structure-aware and task-adaptive.

In our work, because our framework must jointly reason over (*i*) which relations are informative and *(ii)* which neighbors within those relations carry the strongest evidence, we adopt a Relational Graph Neural Network with Hierarchical Attention (RGHAT) for knowledge graph completion (i.e., medical knowledge expansion) (Z. Zhang et al., 2020a). RGHAT treats an entity's ego-network as a two-level hierarchy: a relation-level attention module selects the most salient relation types for the central entity, and an entity-level attention module then weights neighbors within each selected relation before aggregating messages. This yields relation-aware message passing and has demonstrated strong link-prediction performance, aligning well with our need to emphasize both relation semantics and neighbor evidence adaptively.

We instantiate the knowledge graph with medical knowledge from the literature (e.g., edges such as depression ← insomnia). Knowledge graph attention then prioritizes entities by their predictive utility for depression detection (e.g., "insomnia" outranking "headache" for our task). The learned entity embeddings and attention weights feed into the classifier to improve prediction performance. Crucially, we embed this within a closed loop: signals emerging from UGC (e.g., mentions of "brain fog," a post-COVID complaint linked to depression (Pfefferbaum & North, 2020) are validated and, when appropriate, incorporated back into the knowledge graph as new entities and relations. The attention model is retrained or adapted on the expanded knowledge graph, allowing the system to re-weight evidence as the knowledge base evolves.

| Table 3. Knowledge Graph Taxonomy and the Application in Our Work | | |
|---|---|---|
| **KGAT Taxonomy** | **Application in Our Work** | **Our Contribution** |
| Knowledge Acquisition | Newly identified entities are validated (via confidence thresholds and domain experts) and added to the knowledge graphs, directly contributing to knowledge acquisition. | Automatically expands medical knowledge graphs via depression-related entities on social media data. |
| Representation Learning | - Learn embeddings of medical entities (e.g., "depression") and relationships (e.g., "treatments," "symptoms") using attention mechanisms.<br>- The model dynamically adjusts node/edge importance during training (e.g., prioritizing "insomnia" over "headache" for depression prediction). | Integrates relational graph neural network with hierarchical attention with transformers for context-aware and knowledge-aware embeddings. |
| Temporal Networks | While our work does not explicitly model temporal dynamics, the iterative knowledge graph updates (via new symptom discovery) implicitly capture evolving medical knowledge over time. | Implicitly models evolving knowledge through iterative knowledge graph updates. |
| Knowledge Applications | Uses knowledge graph embeddings to improve depression prediction accuracy. | Develops a novel application for depression prediction and symptom discovery. |

## 3 RESEARCH DESIGN

We propose a Closed-Loop LLM and Knowledge Graph framework for depression detection and medical knowledge expansion from UGC. This framework operates as a continuous loop along two paths (Figure 2). (1) Knowledge-aware depression detection path: the LLM module is designed for depression detection and entity recognition, which integrates entity importance and embeddings derived from the knowledge graph module; and (2) Knowledge refinement and expansion path: refinement and expansion of the knowledge graph based on new entities identified by the LLM module, with human experts in the loop. Through iterative cycles of these two paths, our framework continuously discovers previously unseen depression-related entities and emerging relationships, while improving the model's depression detection performance.

We formulate the research problem as follows: for user $u$, let $p_u = \{p_1,...,p_{n_u}\}$ denote their chronologically ordered posts. The predictive target is $\tilde{y} \in \{1, 0\}$ (at risk of depression vs. not at risk). Let $G$ be a medical knowledge graph with entities and relations. The framework jointly (*i*) learns a classifier $f_\Phi$ that maps UGC and knowledge graph signals to a risk probability $\hat{y}_u$, and (*ii*) discovers triplets $(e_h, r, e_t)$ to expand $G$ into $G'$, where $f_\Psi$ is a triplet scoring function based on the graph embedding.

$$\hat{y}_u = argmax\, p\big(y|f_\Phi(p_{u'}, G)\big) \quad (1)$$

$$G' = argmax_{G'} \Sigma_{(e_h, r, e_t) \in T_{new}} log(sigmoid(f_\Psi(e_h, r, e_t, G'))) \quad (2)$$

Our design contrasts with alternative approaches that run prediction and knowledge discovery as separate modules in an iterative pipeline. In such decoupled setups, the two objectives remain only loosely linked, with no principled mechanism to ensure that task-utility signals feed back into the next training cycle. As a result, new entities may be added without regard to whether they truly improve prediction, introducing noise and redundancy, and the system lacks gradient-level guidance connecting knowledge graph updates to downstream performance. To address these limitations, we couple objectives through a joint loss that integrates prediction loss, a utility-regularized graph expansion term that prioritizes high-value entities, and a stability term that discourages volatile updates.

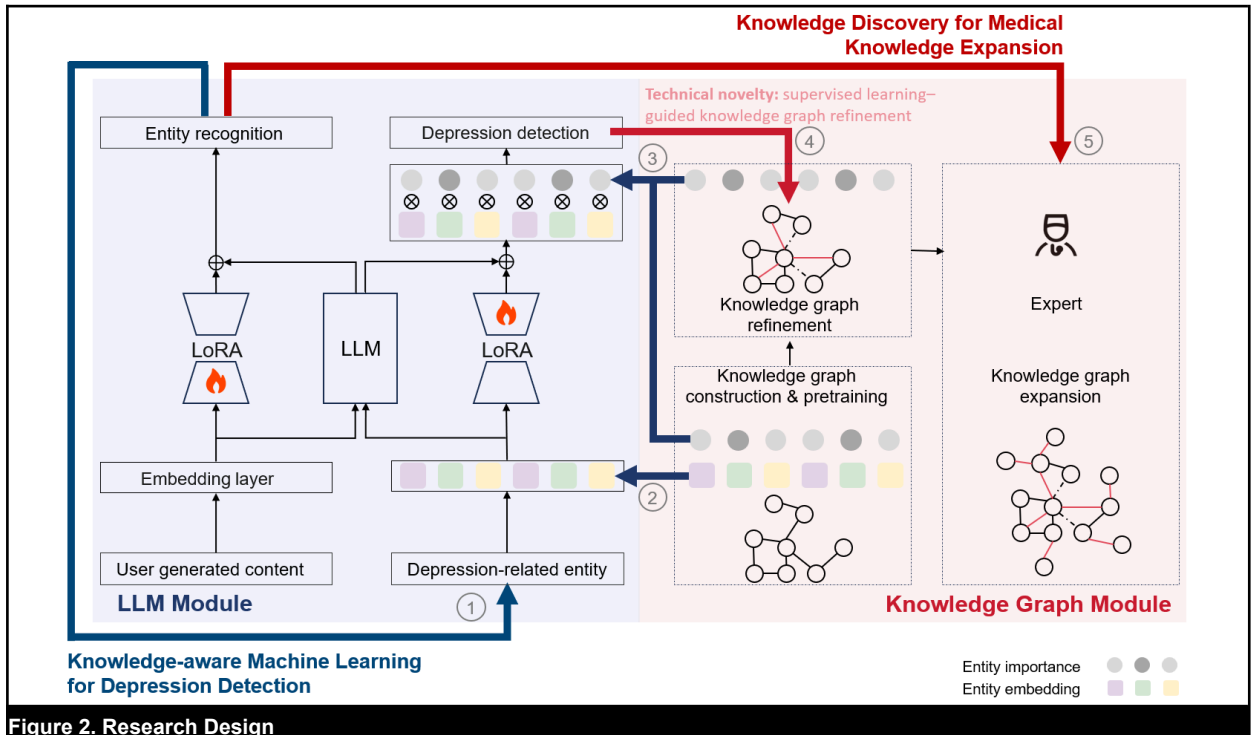

**Figure 2. Research Design**
**Note**: ① Extraction of depression-related entities from user-generated content (UGC) for depression detection (Section 3.1.1).
② The knowledge graph provides entity embeddings (and updated embeddings as the knowledge graph evolves) to support depression detection (Section 3.1.2 and Section 3.2.2).
③ The knowledge graph provides entity importance (and updated importance as the knowledge graph evolves) to support depression detection (Section 3.2.1 and Section 3.2.2).
④ Depression detection (supervised learning) guides the refinement of the knowledge graph (Section 3.2.2).
⑤ Entity recognition contributes to medical knowledge expansion with experts in the loop (Section 3.2.2).

### 3.1 Two Modules: LLM Module and Knowledge Graph Module

We introduce the two backbone modules used to power the two pathways of the closed-loop framework in this subsection. In section 3.2, we will explain the closed-loop framework.

**3.1.1 The LLM Module**

The *LLM module* encompasses two joint learning tasks: (1) the extraction of depression-related entities from UGC, and (2) the prediction task for depression detection facilitated by the knowledge graph module. This section introduces the extraction of depression-related entities within the *LLM module*. The depression detection task, which is supported by the *Knowledge Graph module*, will be introduced in section 3.2.1.

Within the *LLM module*, the entity recognition task serves two purposes. (*i*) We use recognized depression-related entities as inputs for the depression detection task. Therefore, the entity extraction process, which identifies only those entities relevant to depression diagnosis from highly noisy data, can enhance the performance of the depression detection task. (*ii*) More accurate depression detection helps identify the most informative depression-related entities from UGC. The advantage of fine-tuning the LLM within this module using a joint learning design lies in its ability to improve generalization to new, unseen data by capturing shared linguistic patterns across the two tasks (Standley et al., 2020). The shared parameters serve as a form of regularization, helping to mitigate overfitting to noise specific to any single task (Standley et al., 2020). Moreover, within the closed-loop framework, that is, between the *LLM module* and the *Knowledge Graph module*, the depression-related entity recognition task also drives knowledge completion: depression-related entities extracted from UGC that are absent from existing knowledge bases are systematically distilled, validated, and incorporated into the knowledge graph (see section 3.2.2).

To extract depression-related entities, we fine-tune a pretrained LLM to recognize depression-related entities in UGC using Begin, Inside, Outside (BIO) tagging. Three layers of the LLM module are activated: embedding layer, backbone layer, and entity recognition layer.

Let $p_u = \{p_1, p_2, ..., p_{|u|}\}$ be the social media posts of user $u$, where $|u|$ is the total number of posts. For each post $p_i$, let $t_i = (t_{1i}, t_{2i}, ..., t_{ki})$ be the tokens in the post. The set of tokens for user $u$ is denoted as $t_u = \{t_1, t_2, ..., t_{|u|}\}$ and it serves as the input to the LLM for entity recognition. First, the embedding layer maps the input tokens $t_u$ into vector representations $e_u$ using the LLM's pretrained embeddings. The $e_u$ then serve as input of the LLM backbone layer. Then, we fine-tune the LLM backbone layer using the LoRA approach, considering its advantage in multiple aspects. First, LoRA has demonstrated its adequate accuracy and computational efficiency in various LLM fine-tuning tasks (Hu et al., 2021). Second, the LoRA module is lightweight so that different LoRA adapters can be trained for different tasks and dynamically loaded (Feng et al., 2024). This is particularly suitable for our joint learning design. Third, the low-rank constraint acts as implicit regularization and avoids the overfitting issue over a small training dataset (Hu et al., 2021). In our joint learning design, we implement LoRA using a layer-wise task-specific manner so that lower layers and higher layers can focus on task-discriminative features. We fine-tune the parameters of lower layers in LoRA while freezing the parameters of top layers in LoRA in the entity recognition task. In the depression detection task, the parameters of higher layers in LoRA are tuned while the lower layers in LoRA are frozen. Lastly, the entity recognition layer takes the output of the LLM backbone, $e_{u\_ner} \in R^{L_1 \times d_1}$, and input it into a two-layer multi-layer perceptron that maps it to the probabilities for the BIO label class:

$$\tilde{z} = softmax(W_2 Dropout(ReLU(W_1 e_{u\_ner} + b_1)) + b_2) \quad (3)$$

The cross-entropy loss is adopted for the entity recognition task:

$$Loss_{ner} = -\frac{1}{N}\sum_{i=1}^{N}\sum_{c=1}^{3} z_{ic} \cdot log(\widetilde{z}_{ic}) \qquad (4)$$

where $N$ is the total number of tokens in users' posts, $\widetilde{z}_{ic}$ represents the $i^{th}$ token's probability for the label class $c$, and $z_{ic}$ is the true class probability.

**3.1.2 The Knowledge Graph Module**

The *Knowledge Graph module* aims to (1) represent existing medical knowledge, (2) quantify the importance of depression-related entities for depression prediction in the LLM module, and (3) incorporate newly discovered medical knowledge (including the entities themselves and their relationships with existing entities) from UGC, facilitated by the *LLM module*.

The knowledge graph is an evolving and extensible system developed in two stages: (1) Construction and pretraining stage: it incorporates existing medical knowledge along with historical labeled UGC from diagnosed depression patients; (2) Refinement and expansion stage: the knowledge graph is refined during the depression detection process, as the *LLM module* identifies both positive and negative cases. Entities that appear in both types of cases, contributing minimally to depression detection, are down-weighted or eliminated, thereby refining the knowledge graph. Moreover, it integrates newly discovered knowledge (both entities and entity relations) from the *LLM module* for knowledge expansion.

In this section, we first introduce the knowledge graph construction and pretraining stage. The refinement and expansion of the knowledge graph, supported by the *LLM module*, are presented in the *Knowledge Expansion Pathway* section.

The knowledge graph construction and pretraining stage involves a knowledge-driven process in which a knowledge graph is constructed to model how various entities related to depression diagnosis (i.e., extracted from medical literature) contribute to depression prediction. This approach is motivated by two key insights: (1) existing medical knowledge regarding depression can enhance the detection of depression from UGC (Chau et al., 2020), and (2) entities associated with depression diagnosis exhibit varying relationships with depression prediction, contributing unequally to its effectiveness (W. Zhang et al., 2024).

**Knowledge Graph Construction.** We first construct a knowledge graph, $G$, following the commonly adopted approach in the mental disorder domain, which builds the graph based on medical domain knowledge (i.e., widely recognized medical literature and clinical depression screening measures) (W. Zhang et al., 2024). At the same time, we emphasize that our knowledge graph is specifically designed to support depression detection from UGC. Therefore, it is tailored to this task; for instance, elements such as genetic factors, which are less likely to appear in UGC, are excluded from our graph. It should also be noted that there is no strict requirement for the initial knowledge graph. When a larger or more comprehensive knowledge graph better suited to the task becomes available, it can be adopted as our starting graph. The knowledge graph is also refined and verified by a psychiatrist at a nationally recognized hospital. Appendix 1[1] details the construction and validation process of the knowledge graph.

The knowledge graph $G$ consists of five categories of depression-related entities: psychological symptoms ($e_{pys\_sym}$), physical symptoms ($e_{phy\_sym}$), life events that may cause or exacerbate depression ($e_{event}$), medications ($e_{med}$), therapies ($e_{therapy}$). These entities are interconnected. For example, medical literature indicates that certain symptoms of depression ($e_{pys\_sym}$ or $e_{phy\_sym}$) can be side effects of specific medications ($e_{med}$), consequences of negative life events ($e_{event}$), or comorbid symptoms of other mental health conditions ($e_{phy\_sym}$).

To model the complex relationships among these entities, we represent them using three types of connections: directed relations (e.g., *Is_subcategory*), undirected co-occurrence relations (e.g., *Psychological symptom co-occurrence*), and undirected complex relations (e.g., *Life event–psychological symptom cause/consequence*). Each triplet $(h, r, t)$ in $G$ denotes the existence of a relation $r$ from the head entity $h$ to the tail entity $t$. The types of entities and relations are summarized in Table 4.

---

[1] Supplementary materials are available here and can also be provided upon request.

| Table 4. Entities and Relations Defined in the Depression Knowledge Graph | |
|---|---|
| **Entities** | |
| • $e_{pys\_sym}$: Psychological symptoms <br> • $e_{phy\_sym}$: Physical symptoms <br> • $e_{event}$: Major life events <br> • $e_{type}$: Entity class, which includes five categories $e_{pys\_sym\_class}$, $e_{med\_class}$, $e_{phy\_sym\_class}$, $e_{therapy\_class}$, $e_{event\_class}$ | • $e_{med}$: Medications <br> • $e_{therapy}$: Therapies |
| **Relations** | |
| **Directed relation** <br> • $r_{subcat}$: Is_subcategory | **Definition:** Assign depression-diagnosis-related entities to the five entity classes. <br> **Example:** <u>dejected mood</u> → <u>psychological symptom</u> <br> - The entity <u>dejected mood</u> belongs to $e_{pys\_sym}$ (i.e., psychological symptom class). |
| **Undirected co-occurrence relation**: <br> • $r_{med\_co}$: Medication co-occurrence <br> • $r_{psy\_co}$: Psychological symptom co-occurrence <br> • $r_{phy\_co}$: Physical symptom co-occurrence <br> • $r_{event\_co}$: Life event co-occurrence <br> • $r_{therapy\_co}$: Therapy co-occurrence | **Definition:** Explicit co-occurrence relationships between entities from the same classes. <br> **Example:** <u>dejected mood</u> ↔ <u>anxious</u> <br> - The entity <u>dejected mood</u> belongs to $e_{pys\_sym}$. <br> - The entity <u>anxious</u> belongs to $e_{pys\_sym}$. <br> - Entities <u>dejected mood</u> and <u>anxious</u> have co-occurrence relations. |
| **Undirected complex relation**: <br> • $r_{phy\_psy\_co}$: Physical and psychological symptom co-occurrence <br> • $r_{life\_psy}$: Life event - psychological symptom cause/consequence <br> • $r_{life\_phy}$: Life event - physical symptom cause/consequence <br> • $r_{therapy\_psy}$: Therapy - psychological symptom treat/side effect <br> • $r_{therapy\_phy}$: Therapy - physical symptom treat/side effect <br> • $r_{life\_therapy}$: Life event - therapy implicit relation <br> • $r_{med\_psy}$: Medication - psychological symptom treat/side effect <br> • $r_{med\_phy}$: Medication - physical symptom treat/side effect <br> • $r_{life\_med}$: Life event - medication implicit relation <br> • $r_{med\_therapy\_co}$: Therapy - medication co-occurrence | **Definition:** Implicit relationships between entities from two different classes. <br> **Example:** <u>poor appetite</u> ↔ <u>paroxetine</u> <br> - The entity <u>poor appetite</u> belongs to $e_{phy\_sym}$ (i.e., physical symptom class). <br> - The entity <u>paroxetine</u> belongs to $e_{med}$ (i.e., medication class). <br> - Entities <u>poor appetite</u> and <u>paroxetine</u> have co-occurrence relations. <br> The relations are undirected because (1) a depressed patient has a <u>poor appetite</u> and is taking <u>paroxetine</u> as treatment; or (2) it can also indicate that taking medicine <u>paroxetine</u> causes the side effect of <u>poor appetite</u>. |

Notably, the constructed knowledge graph $G$ represents existing static medical knowledge of depression (Figure 3 provides an example). By utilizing a knowledge graph, we can effectively capture the highly complex relationships between entities related to depression detection, thereby encoding intricate medical knowledge. Taking the undirected knowledge graph path *fluoxetine ↔ deal with crisis ↔ anxious ↔ aboulomania* as an example, it may imply that the medicine ($e_{med}$) *fluoxetine* is taken during the happening of life event ($e_{event}$) of *dealing with a crisis* and the psychological symptom ($e_{pys\_sym}$) *anxious* is caused by the event ($e_{event}$) *dealing with a crisis*. The psychological symptom ($e_{pys\_sym}$) *anxious* can occur together with the psychological symptom ($e_{pys\_sym}$) *aboulomania*. Another example from the knowledge graph is the path *divorce ↔ anxious ↔ self harm ↔ supportive psychotherapy*. It suggests that the psychological symptom ($e_{pys\_sym}$) *anxious* may be caused by the life event ($e_{event}$) *divorce* and it can cause physical symptoms ($e_{phy\_sym}$) *self harm*. Besides, *supportive psychotherapy* ($e_{therapy}$) can be adopted as a treatment to prevent *self harm*.

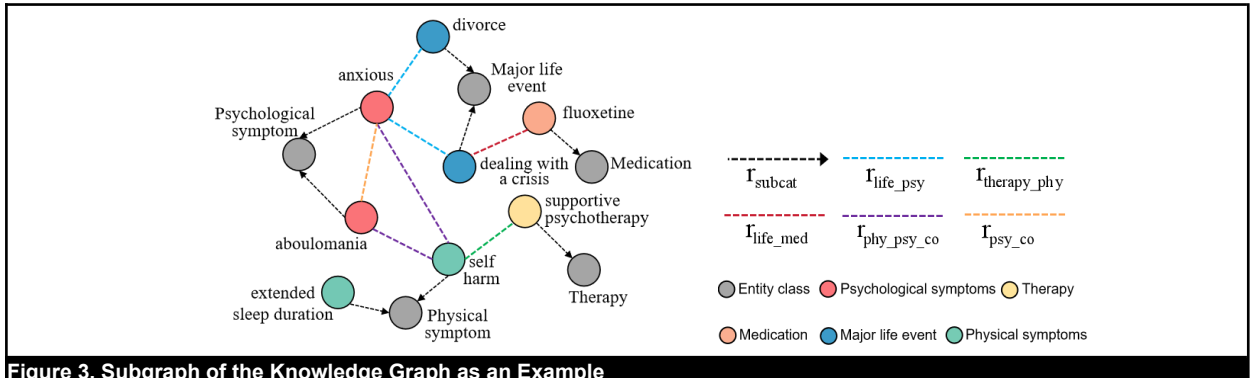

Figure 3. Subgraph of the Knowledge Graph as an Example

**Knowledge Graph Pretraining**. Following the construction of the knowledge graph, we perform supervised graph pretraining using labeled UGC to bridge the critical gap between formal medical knowledge and the specific requirements of depression detection in UGC. The medical knowledge-grounded graph $G$ suffers from three key limitations when applied to this task. (1) Due to the nature of user-generated content, there is a substantial semantic and linguistic gap between such content and formal medical knowledge. Our first aim is to bridge this gap between the informal language used in UGC and the formal language of medical discourse. (2) Entities derived from medical knowledge are typically treated with equal importance, whereas in practice, their frequencies and relevance to depression prediction can vary significantly. Our second objective is to quantify the importance of depression diagnosis-related entities in detecting depression. This is motivated by the fact that depression-diagnosis-related entities extracted from medical literature may exhibit varying degrees of predictive power. For instance, common emotional states such as "sadness" are frequently expressed in UGC by individuals without depression, whereas entities like "recurrent thoughts of death" show significantly stronger associations with a depression diagnosis. In the initial static knowledge graph, all entities are treated as equally important. To optimize the graph, we pretrain the knowledge graph using real-world UGC, relying exclusively on labeled data from users clinically diagnosed with depression. This design allows the originally static entity representations in the knowledge graph to be dynamically updated with associations more directly aligned with the depression prediction task. In doing so, our approach enables the model to overcome the limitations of depending solely on fixed, expert-curated knowledge (Cavalleri et al., 2024). (3) Depression-related entities are interrelated and mutually influential. Our third objective is to quantify the mutual influence among connected entities. To achieve this, we perform entity embedding learning during the training of the knowledge graph, which generates vector representations of the entities. The learned entity embeddings will subsequently serve as input to the *LLM module* for depression detection. This is because, in a knowledge graph, neighboring entities and relations also contribute unevenly to the learning of a focal entity's embedding. Taking the knowledge graph in Figure 3 as an example, "extended sleep duration" and "self harm" are both neighbor entities of $e_{phy\_sym}$ (physical symptoms) via the relation of $r_{subcat}$ (Is_subcategory.) Nevertheless, "self harm" contributes more information to the representation of patients' physical symptoms and to depression detection. Similarly, the $r_{med\_co}$ (medication co-occurrence) relation may contain more medical information than other co-occurrence relations. Knowledge graphs can further model the strength of various relationships and quantify the mutual influence between entities within the complex network. The resulting entity embeddings encapsulate information about both the entity itself and its related entities and relationships. These embeddings, enriched with medical knowledge, can then serve as input to the *LLM module* for the depression detection task.

Specifically, we employ a relational graph neural network (Zhang et al., 2020) with a hierarchical attention mechanism to learn entity embeddings from the knowledge graph $G$.

Furthermore, leveraging the trained relational graph, we estimate the importance score of each entity for depression prediction by calculating the transition probability from each entity node to the "depression" entity along various paths within the graph $G$. Therefore, through this pretraining process, we update both the relationships to depression prediction (represented by attention scores) and the embeddings of entities within the knowledge graph. The pretraining of the knowledge graph occurs before the depression detection function of the *LLM module* begins to generate results. The updated attention scores and entity embeddings can, in turn, serve as weights and inputs, respectively, to enhance the LLM module's depression detection performance (Eqs. 15 and 16).

Within the UGC, entities related to depression diagnosis are identified using the entity recognition function of the *LLM module*. Then, we apply cosine similarity to align existing medical terms with terms extracted from UGC. Next, entities co-occurring with the same post are paired and connected using the predefined entity relations (Table 4), forming a set of triplets as positive samples for entity embedding learning. With these positive triplet examples $(h, r, t)$, we learn graph embeddings (i.e., using ConvE (Dettmers et al., 2018)) and compute the triplet score:

$$f(h, r, t) = ReLU(vec(ReLU([\bar{h}; \bar{r}] * w))Q)t \tag{5}$$

where $*$ is the convolution operator. $vec(\cdot)$ denotes a vectorization function, and $Q$ is a weight matrix. Then, we define the loss function for graph training as:

$$Loss_{KG} = - \log\sigma(f(h, r, t)) - \sum_{i}^{N}(\log(\sigma(-f(h', r, t')))) \tag{6}$$

where $(h', r, t')$ denotes the negative triplets generated by randomly replacing either $h$ or $t$. $N$ represents the number of negative triplets for each positive triplet.

Let $F = (F_1, F_2, ..., F_{|u|})$ be the diagnosis-related entities for user $u$, and $|u|$ represents the total number of extracted entities from the user's posts. The pretrained knowledge graph $G$ can produce the graph embeddings of these entities, denoted as $\tilde{e}_{graph}$, which will serve as the input for the *LLM module* (Eq. 15).

### 3.2 Closed-Loop Framework with Two Pathways

The closed-loop framework consists of two pathways: a knowledge-aware depression detection pathway and a knowledge discovery pathway for medical knowledge expansion. Iterations of these two pathways form the closed-loop framework. Both pathways rely on the interaction between the *LLM module* and the *Knowledge Graph module* introduced in section 3.1.

#### 3.2.1 The Knowledge-Aware Depression Detection Pathway

The knowledge-aware depression detection pathway activates the *Knowledge Graph module,* serving as a knowledge base by generating entity embeddings and importance values for the *LLM module*. Within the *LLM module*, the classification task of the joint learning process is activated to achieve the ultimate objective of depression detection in this pathway.

**Knowledge Graph Module: Entity Importance via Path Probabilities.** As mentioned earlier, depression-related entities may exhibit varying degrees of predictive power. We utilize knowledge in pretrained $G$ to estimate the predictive importance of entities. Since a knowledge graph can model the strength of various relationships and quantify the mutual influence between entities, entities with higher predictive power shall exert a stronger influence on the factor class entity node $e_{type}$, which embeds collective knowledge of each entity class (e.g., $e_{pys\_sym\_class}$ connect with all specific psychological symptom entities). We estimate the entity importance by assessing the transition probability from a target entity to each entity class node $e_{type}$, and subsequently the *depression* entity (predictive power to depression detection).

In our context, when calculating the transition probability, it is essential to account for two types of relationships between different neighboring entities: relation-level attention and entity-level attention. (*i*) Relation-level attention is motivated by the observation that different relations vary significantly in their ability to indicate an entity. For example, the relation $r_{life\_psy}$ (life event - psychological symptom cause/consequence) is more informative than $r_{event\_co}$ (life event co-occurrence) when identifying life events that may trigger depression. The former uniquely captures the connection between life events and depression symptoms at the population level, whereas the latter only reflects co-occurrences of life events, which do not necessarily result in depression. The relational-level attention score distinguishes the contribution of different relations to the target entity. (*ii*) Entity-level attention, on the other hand, is motivated by the intuition that neighboring entities under the same relation may also differ in their importance. For instance, the relation $r_{subcat}$ (Is_subcategory) links $e_{pys\_sym}$ (Psychological symptom) to a set of symptoms. Among these, the symptom *thoughts of self-harm* may be more indicative of depression than symptoms such as *sadness, feeling down, or emptiness,* since many people may occasionally experience such negative emotions. The entity-level attention score evaluates the contribution of different neighbor entities connected with the target entity.

We denote the set of neighbor relations of entity $h$ as $N_h$. For a relation $r \in N_h$, its relational-level attention score $\alpha_{h,r}$ is computed as

$$a_{h,r} = W_1 [h||v_r] \quad (7)$$

$$\alpha_{h,r} = softmax_r(a_{h,r}) = \frac{exp(\sigma(p \cdot a_{h,r}))}{\Sigma_{r' \in N_h} exp(\sigma(p \cdot a_{h,r'}))} \quad (8)$$

where $h \in R^d$ is the embedding of entity $h$ and $d$ is the embedding size. $v_r \in R^d$, $W_1 \in R^{d \times 2d}$, and $p \in R^d$ are trainable parameters. "||" represents the concatenation operation. For neighbor entities of $h$, entities under the same relation $r$ are denoted as $N_{h,r}$. The entity-level attention score $\beta_{r,t}$ is computed as

$$b_{h,r,t} = W_2 [a_{h,r}||t] \quad (9)$$

$$\beta_{r,t} = softmax_t(b_{h,r,t}) = \frac{exp(\sigma(q \cdot b_{h,r,t}))}{\Sigma_{t' \in N_{h,r}} exp(\sigma(q \cdot b_{h,r,t'}))} \quad (10)$$

where $t \in R^d$ is the embedding of the tail entity $t$ under the relationship of $r$. $W_2 \in R^{d \times 2d}$ and $q \in R^d$ are trainable parameters. The obtained $\beta_{r,t}$ represents the attention score of $t$ when representing its head entity $h$.

Once both relational-level and entity-level attention are obtained, their integration is used to evaluate the contribution of a triplet $(h, r, t)$ in characterizing the head entity $h$. Specifically, the relation-level attention score $\alpha_{h,r}$ and the entity-level attention score $\beta_{r,t}$ are combined to compute the triple-level attention score:

$$\gamma_{h,r,t} = \alpha_{h,r} \cdot \beta_{r,t} \tag{11}$$

$\gamma_{h,r,t}$ serves to identify and distinguish the most informative neighboring entities and relations.

The transition score for each pair of entity neighbors is then computed as:

$$r = exp(\gamma_{h,r,t} \times \sqrt{N_{h,t}}) \times log(c_t + 1) \tag{12}$$

where $N_{h,t}$ represents the number of tail entities connected with entity $h$, $\sqrt{N_{h,t}}$ rescales the triple-level attention score to mitigate the attenuation effect caused by larger tail entity numbers after Softmax transformation, $c_t$ represents the number of tail entities from the head entity $h$. For a transition path $e_1 \rightarrow e_2 \rightarrow ... e_n \rightarrow e_{type} \rightarrow depression$ with the transition scores of $r_1 \rightarrow r_2 \rightarrow ... r_n$, the transition probability from $e_1$ to $e_{type}$ is computed as $r_{path} = \prod_{i=1}^{n} r_i$. The transition score for $e_{type} \rightarrow depression$ represents the importance of the entity class (e.g. psychological symptom class) and is determined according to class entity frequency and existing medical knowledge.

Our next objective is twofold: (1) to identify the most probable transition path among multiple possible paths from a target entity $e_1$ to its entity type ($e_{type}$), and (2) to estimate entity importance. To operationalize this process, we propose a new approach that employs the Monte Carlo Tree Search (MCTS) method (Steven et al., 2017) to compute the transition probabilities from the target entity $e_1$ to $e_{type}$ via different paths in the knowledge graph (detailed algorithm is provided in Appendix 3). The reward of action in MCTS is defined as the transition probability in order to find the most probable transition path. The largest transition probability value $r_{path}$ is adopted as the entity importance.

For a depression-related entity $F_i$ generated from UGC, the top $M$ most similar entities in $G$ are selected using a similarity measure:

$$sim(F_i, G_j) = \frac{e_{F_i} e_{G_j}^T}{\|e_{F_i}\| \|e_{G_j}\|} \tag{13}$$

where $e_{F_i}$ is the embedding of entity $F_i$, $e_{G_j}$ is the embedding of entity $G_j$ in the knowledge graph $G$. Then, the predictive importance score of $F_i$ is computed as:

$$w_{Fi} = \sum_{m=1}^{M} sim(F_i, G_{j,m}) \cdot r_{path,m} \tag{14}$$

where $G_{j,m}$ belongs to the top $M$ most similar entities in $G$ to $F_i$.

The importance weights of depression-related entities $w_F = (w_{F1}, w_{F2}, ..., w_{F|u|})$ and their graph embeddings $\tilde{F}_{G\_u} = (\tilde{F}_1, ..., \tilde{F}_i, ..., \tilde{F}_{|u|})$ serve as input for the *LLM module* (Eq. 15-16).

**LLM module: Knowledge-aware Depression Detection.** For the depression detection task, we use the previously identified depression-related entities $e_{u\_ner}$ (embeddings from the *LLM module*'s depression-related entity recognition task), retrieve their knowledge graph embeddings $e_{graph}$ and importance scores $w_{u\_dep}$, and predict the focal user's risk of depression. Two layers in the *LLM module* are activated: the backbone layer and the depression classification layer.

The input for the LLM backbone layer is the residual fusion of the two entity embeddings:

$$e_{u\_ent} = e_{u\_ner} + Dropout(ReLU(W_3 [e_{u\_ner}, e_{graph}])) \tag{15}$$

where [,] represents the concatenation operation, $W_3 \in R^{(d_1+d_2) \times d_1}$ is a projection matrix, and $d_1$ and $d_2$ represent the dimensions of $e_{u\_ner}$ and $e_{graph}$, respectively. First, the backbone layer of LLM is fine-tuned using LoRA, with adaptation applied to its higher layers. Second, let the $e_{u\_dep} = (e_{u\_dep1}, e_{u\_dep2}, ..., e_{|u\_dep|})$ be the output embeddings from the LLM backbone layer.

We account for the unequal contribution of each entity for depression detection and integrate the entity importance values (as defined in the *Knowledge Graph Module: Entity Importance via Path Probabilities* section) with $e_{u\_dep}$ to optimize the information of the entities. Let $w_{u\_dep} = (w_{u\_dep1}, w_{u\_dep2}, ..., w_{|u\_dep|})$ be the importance weights obtained from the entity importance estimation step (Eq. 14), the importance-adjusted entity embeddings is obtained by:

$$e_{u\_dep}' = [e_{u\_dep1} \cdot w_{u\_dep1}, e_{u\_dep2} \cdot w_{u\_dep2}, ..., e_{|u\_dep|} \cdot w_{|u\_dep|}] \quad (16)$$

where [,] is the concatenation operation. Lastly, the depression prediction layer input $e_{u\_dep}'$ into a two-layer multi-layer perceptron that maps it to the probabilities for the depression class:

$$\tilde{y} = sigmoid(W_5 Dropout(ReLU(W_4 e_{u\_dep}' + b_3)) + b_4) \quad (17)$$

The cross-entropy loss is adopted for the depression prediction task:

$$Loss_{dep} = -\frac{1}{|U|}\sum_{i=1}^{|U|}(y_i \cdot log(\tilde{y}_i) + (1 - y_i) \cdot log(\tilde{y}_i)) \quad (18)$$

where $\tilde{y}_i$ represents the prediction result for $i^{th}$ user and $y_i$ represents the user's true depression class, $|U|$ represents the total number of users.

The LLM module's joint objective is

$$Loss_{LLM} = Loss_{ner} + \lambda Loss_{dep} \quad (19)$$

with $\lambda > 0$ balancing entity recognition and depression detection. We alternatively optimize $Loss_{ner}$ and $Loss_{dep}$. The two tasks share the LLM backbone. LoRA adapters are applied layer-wise to stabilize multitask fine-tuning without overfitting.

### 3.2.2 The Knowledge Expansion Pathway

The closed-loop framework incorporates a second path focused on knowledge discovery for medical knowledge expansion. This path involves (*i*) refining the knowledge graph through depression depression in the *LLM module* and (*ii*) expanding the knowledge graph by identifying new depression-related entities through the *LLM module*.

**Knowledge Graph Refinement.** Knowledge graph refinement can be viewed as incremental knowledge learning, where new knowledge is progressively integrated into the graph.

Unlike other knowledge graph refinement processes, we propose a novel refinement approach driven by the depression detection process in the *LLM module*. This process is embedded within our closed-loop system, enabling the knowledge refinement to be achieved through iterative feedback between the *LLM module* and the *Knowledge Graph module*.

As the *LLM module* processes new UGC, it yields (*i*) positive triplets co-mentioned by depressed users and (*ii*) negative triplets co-mentioned by non-depressed users. A critical challenge is to mitigate bias arising from false positives or false negatives in the identification of depression-related entities. For instance, the triplets $(h, r, t)_{positive}$ may overlap with triplets in $(h, r, t)_{negative}$ as many users, either depressed or non-depressed, use common words such as "sad" to express their negative emotions, making these entities less indicative in depression detection. These low-quality entities will misguide the convergence of the knowledge graph and cause bias that degrades the quality of knowledge graph embeddings. To mitigate the impact of false positive or false negative triplets, we propose a new penalty weight approach that combines the embedding matching probability with a triplet plausibility score. The design intuition is we reweight each negative triplet's loss using two signals. (*i*) The embedding matching score is the normalized triplet score (Eq. 11), which quantifies how strongly the current embeddings support that triplet. True negatives should have low triplet scores; therefore, when a negative triplet receives a high embedding matching score, we assign a larger penalty so gradient descent pushes its score down more aggressively during training. (*ii*) The negative class plausibility is the sample distribution based probability (Eq. 21) that the triplet truly belongs to the negative class.

E.g., a triplet with plausibility 0.8 is more clearly negative than one with 0.3. We impose a higher penalty to maintain a stronger separation from positives. In summary, the penalty weight for a negative triplet increases with both its embedding matching score and its negative class plausibility, concentrating learning on hard and clearly negative cases.

Given a set of negative triplets $(h, r, t'_1), (h, r, t'_2), \ldots, (h, r, t'_l)$ that conflict with the positive triplets, we first compute the weight of each negative triplet using their embedding matching score:

$$w_s(h, r, t') = \frac{exp(f(h,r,t')/\tau)}{\Sigma_i exp(f(h,r,t'_i)/\tau)} \tag{20}$$

where $f()$ represents the triplet score computed using Eq. 11, and $\tau$ denotes the score temperature coefficient. Then, a triplet plausibility score is calculated based on the sample distribution to represent the confidence of classifying the triplet as either a positive or a negative example. Consider a triplet with $c(h, r, t)$ positive data points and $c(h, r, t')$ negative data points, the plausibility score for the negative class is computed as:

$$p_{(h,r,t')} = \frac{c(h,r,t')}{c(h,r,t)+c(h,r,t')} \tag{21}$$

By combining the embedding matching score and plausibility score, the penalty weight for the negative sample $(h, r, t')$ is:

$$w_c(h, r, t') = p_{(h,r,t')} * w_s(h, r, t') \tag{22}$$

The incremental learning process also adopts the relational graph neural network approach (Zhang et al., 2020). Consequently, negative samples with higher embedding matching probability and higher plausibility score shall receive a higher penalty weight, resulting in a larger penalty from the loss function.

The loss of knowledge graph refinement becomes

$$Loss_{KG-refine} = -\sum_{(h,r,t)\in P} log\sigma(f(h, r, t)) - \sum_{(h,r,t)\in N} (w_c(h', r, t')log(\sigma(-f(h', r, t'))) \tag{23}$$

where *P* denotes the set of positive samples and *N* denotes the set of negative samples. The refinement process updates both embeddings and attention weights, resulting in improved entity embeddings and recalibrated importance scores, which are then fed back into the *LLM module* to enhance depression detection (Eqs. 15 and 16). This step happens after the LLM-based depression detection module converges and provides sufficiently accurate results.

**Knowledge Graph Expansion.** After iterative training of the *LLM module* and the *Knowledge Graph module* within the closed-loop system, a sufficient volume of UGC has been processed. The *LLM module* has identified new entities and relations that reveal emerging patterns among depressed users on social media. Augmenting the refined knowledge graph using these new entities and relations can narrow down the gap between depression patients' social media behavior and the existing knowledge base.

Specifically, first, we type each new entity into an existing class using the *LLM module* and classify the entities into the pre-defined entity categories. Entities outside the predefined categories are summarized to identify new categories. For example, "black-and-white thinking" and "overthinking" cannot be properly classified into the predefined five categories (psychological symptoms, physical symptoms, life events, medications, therapies), and they imply a new category describing the cognitive patterns of depression patients.

Second, we link the entity to neighboring entities based on co-occurrence and the knowledge graph schema. For a newly detected entity, it can co-occur with: *i*) existing entities in the knowledge graph, *ii*) other newly detected entities, *iii*) no entities in UGC. In scenarios *i*) and *ii*), relations are extracted for the entity pair according to the knowledge graph relation definition (Table 4). In scenario *iii*), the newly detected entity will be linked to the factor class entity ($e_{type}$) as the entity class has been assigned in the previous entity typing step. This step generates new triplets for knowledge graph expansion.

Last, we generate candidate triplets in the form of *(new entity, relation, existing entity)*,

which are then submitted for expert validation. Only expert-approved triplets are added to the knowledge base. We leverage this expert-in-the-loop design in our framework to evaluate the rationality and scientific validity of new triplets. We invited two clinical psychologists from a public hospital to independently screen the triplets and verify: 1) whether the entities in the new triplets are helpful for depression diagnosis? 2) If yes, whether the relations in the new triplets are reasonable? We then assess the consistency of responses from clinical psychologists and retain triplets with consistent feedback for knowledge graph expansion. The augmented knowledge graph then regenerates embeddings and importance scores for the next iteration of depression detection in the *LLM module*. In addition, this new knowledge base incorporates information derived from UGC. These entities capture aspects of depression that are often overlooked in existing clinical studies and scientific literature. As such, they hold potential value for clinical assessment, diagnosis, and ongoing monitoring of depression in patients. Moreover, newly identified entities may also inform future research directions. The algorithm of knowledge graph construction, pretraining, refinement, and expansion is available in the [online repository](online repository).

### 3.2.3 Overall Objective and Training Flow

Our framework alternates between the *LLM module* and the *Knowledge Graph module* updates. At each iteration $k$, the objective is defined as

$$J^{(k)} = Loss_{LLM}^{(k)} + \gamma Loss_{KG-refine}^{(k)} + \mu Loss_{KG}^{(nk)}, \quad \gamma, \mu > 0 \tag{24}$$

where periodic knowledge graph expansion rounds trigger the recomputation of $Loss_{KG}^{(nk)}$ on the enlarged knowledge graph (Note that knowledge graph expansion does not occur at every iteration. The parameter *n* is a hyperparameter that takes a natural number value. After several iterations, $nk$, the knowledge graph is expanded). This closed-loop optimization continuously aligns the *LLM module*'s understanding with an expanding, expert-validated knowledge base, thereby enhancing predictive performance while uncovering clinically meaningful patterns.

This design explicitly addresses the limitations of relying on LLMs alone. Pure LLM approaches struggle to (*i*) dynamically ingest and update structured domain knowledge, (*ii*) accumulate weak yet continuous signals from longitudinal UGC into explicit, testable entity triplets, and (*iii*) quantify how newly surfaced entities relate to established ones. By externalizing memory and structure in an evolving knowledge graph—with measurable importance (Eqs. 7-12), bias-aware refinement (Eqs. 20–23), and expert gating—the framework ensures methodological rigor while remaining adaptive to emergent depression-related entities.

### 3.3 Key Novelties of the Framework

From a design science perspective, our contributions are twofold: (1) We develop a closed-loop framework for depression detection and knowledge expansion. This framework is implemented through a loss function in which the depression-related entity recognition loss, depression detection loss, and knowledge refinement loss are jointly optimized. This means that the system learns all tasks simultaneously rather than sequentially. During training, the model is adjusted to enhance the accuracy of depression detection while improving the structure and precision of the knowledge graph. The framework's parameters are updated to balance these objectives, identifying an optimal point where progress in one task reinforces, rather than compromises, the other. Over time, this joint optimization makes both the prediction task and the knowledge base more intelligent and adaptive. The resulting framework not only effectively identifies depression from UGC but also discovers new depression-related insights, including emerging symptoms, shifts in symptom frequency at the population level, societal events contributing to depression, and previously unrecognized adverse effects of depression treatments. (2) Within this closed-loop framework, we introduce a new relational graph neural network with hierarchical attention with two key enhancements that distinguish it from existing graph neural network models: *i*) Our model accounts for the potential bias introduced by weak positive and negative triplet samples (i.e. duplicated triplets extracted from UGC of depressed users and non-depressed users) by developing a graph refining process incorporating a penalty weight. The proposed penalty weight combines the embedding matching score with the plausibility score, mitigating the impacts of weak positive and negative samples and offering a fine-grained graph learning process. As a result, our approach improves depression detection accuracy and enhances knowledge graph quality. *ii*) We propose a graph-based entity importance estimation approach to quantify the relationship strength between depression-related entities and depression. The entity importance quantifies the transition probability on the most probable transition path from recognized entities to the depression entity.

## 4 EVALUATION

### 4.1 Evaluation Setup

We use eRisk2018 (task 1) and eRisk2022 (task 1) as the testbed (Losada et al., 2018; Parapar et al., 2022). This dataset is commonly used as a benchmark for depression prediction (W. Zhang et al., 2024). It contains 1,866,524 archival Reddit posts (52,247,059 words) generated by 3,870 labeled users, including 457 users with diagnosed depression and 3,143 non-depression users. The summary statistics are shown in Appendix 2.

For the selection of the base LLM, we require an *open-source* model to allow further training. Therefore, the DeepSeek-R1 Distill Llama-8B was selected, as it delivers an exceptional balance between performance and model size. Its relatively small size combined with strong performance makes it an ideal choice for training on domain-specific datasets (DeepSeek-AI et al., 2025). During evaluation, we use 60% of the data for training, 20% for validation, and 20% for testing (we ensure that all posts by the same user are assigned to exactly one split). Performance is evaluated using F1 score, precision, and recall. The reported results represent the mean values across 10 runs. For the deep learning models, we additionally report standard errors to indicate statistical significance. The hyperparameter fine-tuning and Knowledge Graph module pretraining are presented in Appendix 2.

## 4.2 Evaluation of the Closed-loop Framework

To evaluate the capability of the closed-loop framework in continuously detecting depression and expanding the medical knowledge base, we divided the Reddit dataset into nine chronological periods. The first period contains Reddit data from 2006 to 2017 (eRisk2018). The subsequent eight periods cover data from 2018 to 2021 (eRisk2022), each representing a six-month interval. In evaluating the capability of the closed-loop framework, we use the first period as the initial training dataset and progressively incorporate data from each subsequent period into the training set. In this way, the training dataset grows over time, reflecting the temporal progression of the posts and their gradual integration into the closed-loop framework. The testing dataset remains fixed as a randomly selected 12.5% subset of the eRisk2022 data. This design enables us to observe the incremental expansion of the knowledge base at each stage and assess framework performance as new knowledge is discovered and incorporated. This evaluation strategy also simulates a realistic application scenario: an initial dataset is used to train the model, and, as time progresses, additional UGC is continually collected, added to the training process, and used to expand the medical knowledge base.

First, we examine whether medical knowledge expands within our closed-loop framework and whether the growth of this knowledge indeed leads to improved performance in depression detection. After completing the experiments on knowledge graph pretraining, initial training, and the subsequent eight periods, we analyzed the evolution of the knowledge graph. The number of nodes and edges for each period is reported in Table 5. Starting from the pretrained knowledge graph, the number of nodes steadily increases, showing that new depression-related entities from UGC are incrementally incorporated into the graph over time. Likewise, the gradual increase in the number of edges indicates that new relationships between depression-related entities are continuously discovered from UGC. Moreover, we visualize the emergence of new entities in the global knowledge graph and local knowledge graph around the "depressed" entity across time periods (Figure 4). In the initial phase, the entities in the knowledge graph are mainly the treatment and therapy classes, representing existing domain knowledge. New entities (yellow nodes in Figure 4) from the life event, psychological symptom, and physician symptom classes are gradually detected from UGC and added to the knowledge base.

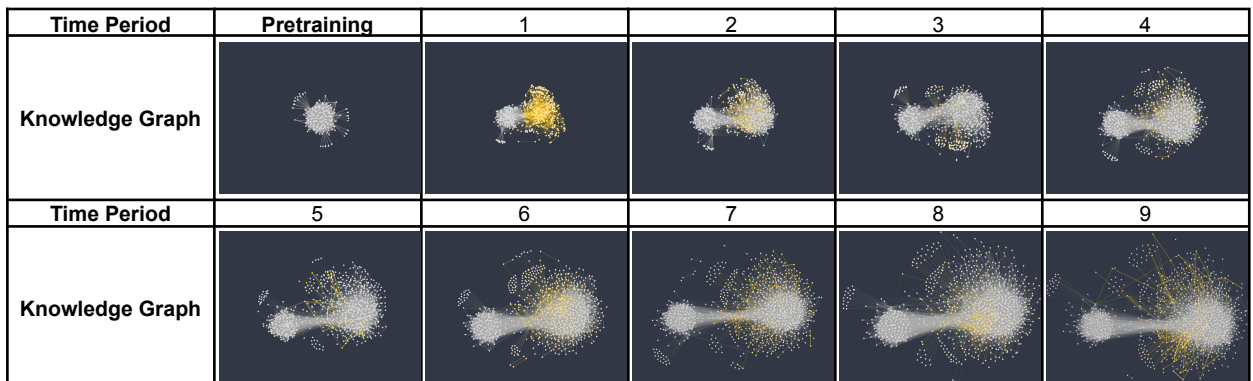

**Figure 4. Visualization of the Knowledge Graph Expansion Process**

The prediction performance of the closed-loop framework for all the time periods is also presented in Table 5. A steady increase in the F1 score is observed over time. The F1 score for the ninth time period is higher than that of the first time period by 5.4%. This upward trend demonstrates the contribution of newly identified depression knowledge and the advantage of designing a closed loop in knowledge-aware depression detection models. The knowledge expansion module allows the model to learn on top of what is already known in the medical field and adapt to the specific task context. Context-specific knowledge is effectively exploited in the subsequent depression detection tasks. By contrast, the "*Performance with Increasing Training Data*" column in Table 5 presents the results of the base model with additional training data. The F1 score shows only a marginal improvement despite the increase in training data, suggesting that the performance gain achieved by our proposed framework can be attributed to knowledge expansion rather than merely to the growth of the training dataset.

**Table 5. Statistics on Knowledge Graph Expansion and Prediction Performance Across All Time Periods**

| Time Period | Number of nodes | Number of edges | Performance with Increasing Training Data | Ours: Closed-Loop Framework with Knowledge Graph Expansion | | |
|---|---|---|---|---|---|---|
| | | | F1 | F1 | Precision | Recall |
| Knowledge Graph Pretraining | 249 | 10,930 | – | – | – | – |
| 1 | 596 | 13,339 | 0.816 ± 0.010 | 0.816 ± 0.010 | 0.846 ± 0.022 | 0.795 ± 0.016 |
| 2 | 692 | 18,352 | 0.819 ± 0.003 | 0.831 ± 0.006 | 0.888 ± 0.021 | 0.796 ± 0.008 |
| 3 | 768 | 20,297 | 0.820 ± 0.009 | 0.835 ± 0.004 | 0.891 ± 0.031 | 0.801 ± 0.014 |
| 4 | 842 | 25,446 | 0.823 ± 0.009 | 0.837 ± 0.002 | 0.904 ± 0.028 | 0.798 ± 0.013 |
| 5 | 857 | 27,657 | 0.823 ± 0.007 | 0.839 ± 0.006 | 0.891 ± 0.021 | 0.806 ± 0.015 |
| 6 | 975 | 34,775 | 0.825 ± 0.008 | 0.840 ± 0.005 | 0.888 ± 0.021 | 0.809 ± 0.013 |
| 7 | 1,043 | 38,384 | 0.824 ± 0.007 | 0.853 ± 0.006 | 0.900 ± 0.022 | 0.822 ± 0.009 |
| 8 | 1,134 | 46,418 | 0.823 ± 0.008 | 0.858 ± 0.005 | 0.896 ± 0.015 | 0.830 ± 0.008 |
| 9 | 1,189 | 53,636 | 0.825 ± 0.008 | 0.860 ± 0.005 | 0.906 ± 0.014 | 0.829 ± 0.005 |

Second, we aim to demonstrate that the improvement in depression detection within our closed-loop framework stems from the expanded knowledge base derived from UGC. We further illustrate this enhancement with patient-level examples. The increasing Recall scores in Table 5 indicate that our framework identifies more depressed patients as its knowledge accumulates. There are two possible reasons. (*i*) The first reason is that our closed-loop framework can identify more depression-related entities. Patient #5031 exemplifies this case. In the early stages, the entities recognized from user posts were limited, including "expensive," "beat," "ill," and "abandoned," resulting in relatively weak signals for detection. Consequently, the depression detection component failed to classify this patient as depressed. In later stages, as the knowledge graph expanded, the framework identified additional depression-related entities from Patient #5031, such as "disrespectful," "upset," "trouble," "bugs," "anxiety," and "fight," which provided stronger evidence for depression detection. Our model ultimately classified this patient as depressed, with corresponding improvements in recall and F1 score. (*ii*) The second reason is that, as the knowledge graph expands, our closed-loop framework can continuously refine the importance scores of depression-related entities. Figure 6 illustrates how the importance scores of several sample entities evolve over time. For instance, the importance score of "nightmare" increases from 0.393 to 0.620 through the iterative closed-loop learning process. This notable increase suggests that, based on the new knowledge learned from UGC, when a user's posts contain "nightmare," the model is more likely to identify them as exhibiting signs of depression. Patient #3362 is one such example: "I have dreams about getting murdered or murdering people. I fight with my family - they're very realistic. I get messed up, and have all sorts of random, vivid nightmares." As the knowledge graph feedback refined the importance of related entities, our framework correctly classified this patient as depressed in the later stage, thereby improving our framework's recall and F1. Another contrasting example is the entity importance score for "awkward," which decreases from 0.422 to 0.163 across the time periods, suggesting that it is a weak indicator of depression. Patient #3880 falls into this category: "Why does puberty cause all those awkward side effects, like changes in voice and acne?" Based on the knowledge graph feedback regarding the importance score of "awkward," our model later classified this patient as not having depression. The adjustment in entity importance scores amplifies useful signals for depression prediction and improves both the precision and F1 score of our framework.

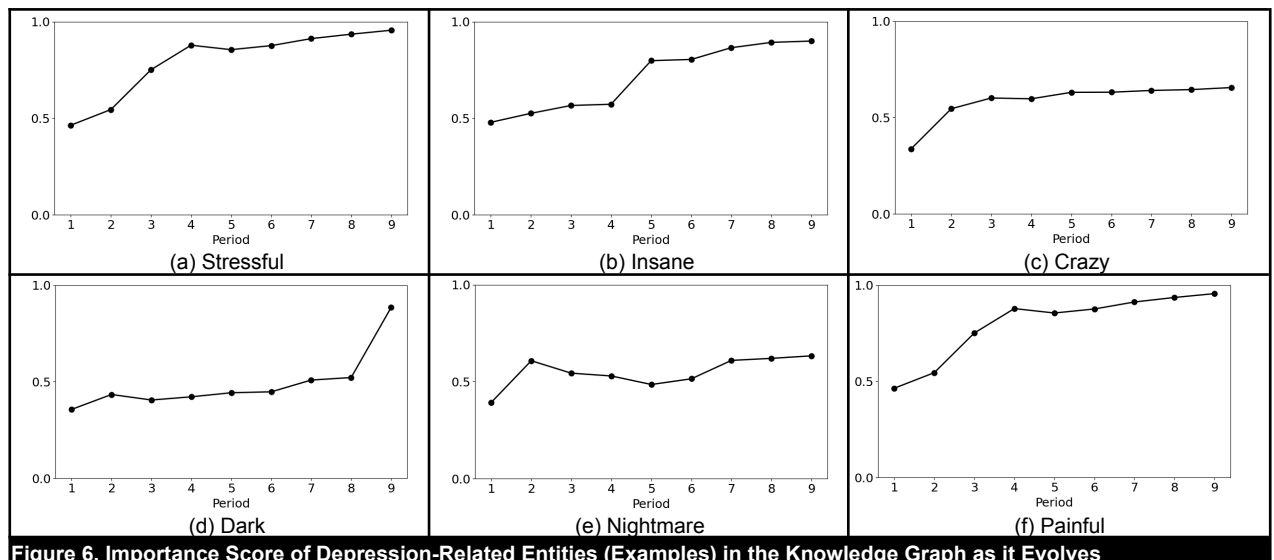

**Figure 6. Importance Score of Depression-Related Entities (Examples) in the Knowledge Graph as it Evolves**

Furthermore, we examine whether our framework, when trained on the most recent time period and incorporating knowledge accumulated across all prior periods, can enable early predication, a task that our original framework was unable to perform reliably. For example, Patient #6326 began to exhibit signs of depression in time period 3, as reflected in the post: "I recently quit my job because of stress, at my doctor's recommendation …" The framework trained only on data up to time period 3 incorrectly classified this patient as non-depressed. In contrast, the framework trained on data up to the most recent time period correctly identified the patient based on the same input. This finding suggests that continuously integrating knowledge across time periods enhances the framework's capacity to produce accurate early predictions.

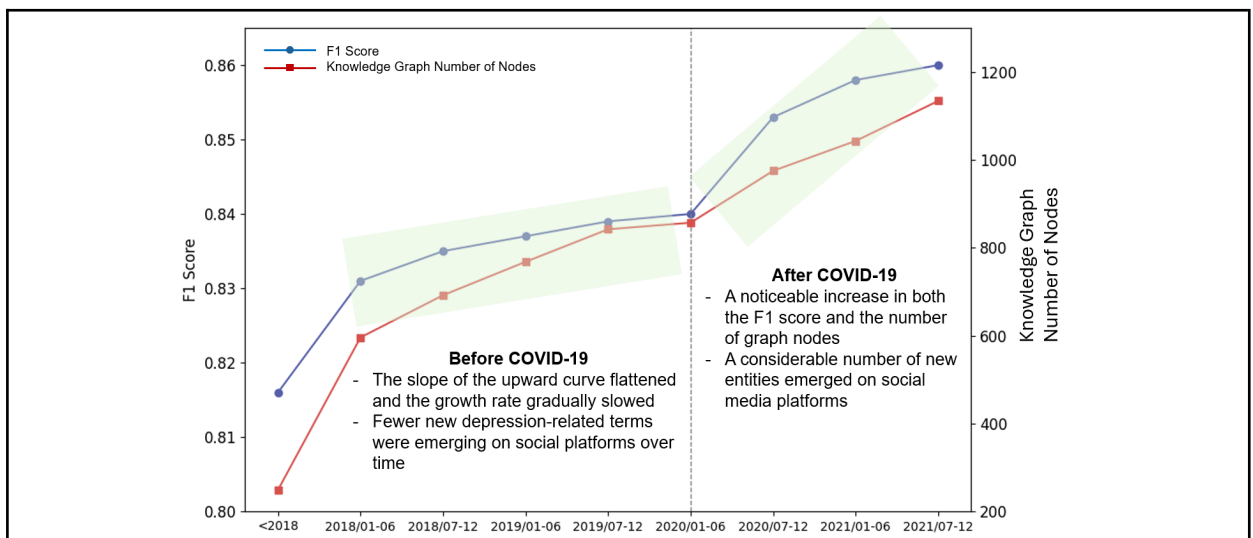
**Figure 7. Alignment of Newly Discovered Knowledge with Performance Gains and Its Connection to Real-World Events.**

We further analyze the alignment between newly discovered knowledge and depression detection performance gains, as well as its connection to real-world events. We visualize the model's F1 score and number of nodes in the knowledge graph for all the time periods in Figure 7. The F1 score increases gradually, together with the number of graph nodes over time. For the time periods before Jan 2020, the slope of the upward curve has flattened, and the growth rate has gradually slowed, indicating that there are fewer new depression-related entities emerging in UGC over time. However, there is a burst of number of graph nodes and F1 score for the period from Jan 2020 to Jan 2021, indicating that a considerable number of new entities emerge in UGC. We examine the newly detected entities (e.g., "flu," "infectious," "isolation," "starve," "plague," "deteriorate," "pressure," "lifespan," "cough," "lock," "devastating," "respiratory," "mortality," "faint") and observe that these emerging depression related entities correlate temporally with COVID-19 pandemic discourse (2020-2021). This indicates that the various negative impacts of the COVID-19 pandemic on public health and daily life are extensively discussed on social platforms, which in turn exerted considerable influence on population mental health (Pfefferbaum & North, 2020).

### 4.3 Evaluation of Expanded Medical Knowledge

To ensure clinical relevance, we conduct an expert-in-the-loop validation and consolidation of the newly discovered medical knowledge, including depression-related entities and their relationships represented as entity triplets, following the final iteration of the learning process. First, we invited three mental health professionals (mean clinical experience >5 years) to evaluate the clinical relevance and validity of entities. The evaluation is based on DSM-5 diagnostic criteria (APA, 2022) using a 5-point Likert scale (1 = irrelevant, 5 = highly relevant). Second, they evaluated the contextual appropriateness of the newly identified triplets using a 5-point Likert scale (1 = irrelevant, 5 = highly relevant). Each entity and triplet is evaluated by at least two professionals, resulting in an inter-rater reliability score (Cohen's Kappa) of 0.82. Entities and triplets with average scores above 4 in both clinical relevance/validity and contextual appropriateness are retained as new depression knowledge. Third, the professionals categorize the entities into the five predefined classes (Table 4). For entities that cannot be classified into these existing categories, the professionals identify and summarize new categories (e.g., cognitive patterns). Finally, all newly detected triplets were reviewed by medical professionals to resolve discrepancies and establish a consensus through iterative discussion. This validation process ensures the clinical validity and soundness of the identified entities and their associated relationships, improving the clinical reliability of subsequent findings.

To illustrate the new knowledge learned by the framework, we provide a few representative examples of newly identified knowledge triplets, consisting of specific entities and their relationships with existing entities (Figure 5). There are three types of newly identified knowledge triplets. (*i*) The first type is triplets with "Is_subcategory" relations (Table 4), representing new depression-related entities of the five predefined entity classes. For example, new entities detected for the life event class in the 1st time period include "heroin," "cheat," "job," "debt," "discrimination," "cancer," "defeat," "crisis," "prison," etc. These entities are all connected with life event ($e_{type}$) node in the knowledge graph via the "Is_subcategory" relation (Figure 5a and 5b). (*ii*) The second type is the triplets with an undirected co-occurrence relation. For example, triplets with psychological symptom co-occurrence relations include "angry" ($e_{pys\_sym}$) and "numb" ($e_{pys\_sym}$), "panic" ($e_{pys\_sym}$) and "shock" ($e_{pys\_sym}$), "upset" ($e_{pys\_sym}$) and "irrational" ($e_{pys\_sym}$), "confuse" ($e_{pys\_sym}$) and "lonely" ($e_{pys\_sym}$), etc (Figure 5c and 5d). Discovering the co-occurrence of psychological symptoms provides stronger evidence for depression detection than examining a single symptom. For example, a symptom such as "fatigue," which is listed in existing medical knowledge bases and diagnostic scales, can arise from multiple causes and is not unique to depression. By analyzing UGC, we find that real-world social media users suggest that psychological symptoms rarely occur independently; rather, they tend to emerge as interrelated clusters, which serve as strong indicators for depression detection. UGC triplets with major life event occurrence relations include "unemployed" ($e_{event}$) and "debt" ($e_{event}$), "epidemic" ($e_{event}$) and "death" ($e_{event}$), "poverty" ($e_{event}$) and "welfare" ($e_{event}$), "censorship" ($e_{event}$) and "ban" ($e_{event}$), etc (Figure 5e and 5f). These co-occurring major life events suggest that events with negative effects on individuals' physical and psychological well-being may not occur in isolation but rather form event chains that exert a deeper impact on mental health. The co-occurrence of these life events can assist medical professionals in identifying factors that may trigger or exacerbate depression, as well as in evaluating its severity. (*iii*) The third type is triplets with an undirected complex relation, such as physical and psychological symptom co-occurrence ($r_{phy\_psy\_co}$). Triplets indicating the co-occurrence of physical and psychological symptoms include "stomach ache" ($e_{phy\_sym}$) and "crazy" ($e_{pys\_sym}$), "choke" ($e_{phy\_sym}$) and "crazy" ($e_{pys\_sym}$), "alcoholic" ($e_{phy\_sym}$) and "anxiety" ($e_{pys\_sym}$), etc (Figure 5g and 5h). The co-occurrence of these entities suggests that physical and psychological symptoms tend to co-occur. An alternative explanation is that people tend to use physical vocabulary to describe complex psychological states to make their feelings more concrete (Lakoff & Johnson, 2024). In addition, these triplets support the established symptom clusters defined in psychiatry and medicine. For instance, the association between gastrointestinal dysfunction (such as "stomach ache") and depression constitutes a central pillar of gut–brain axis research (Cryan & Dinan, 2012).

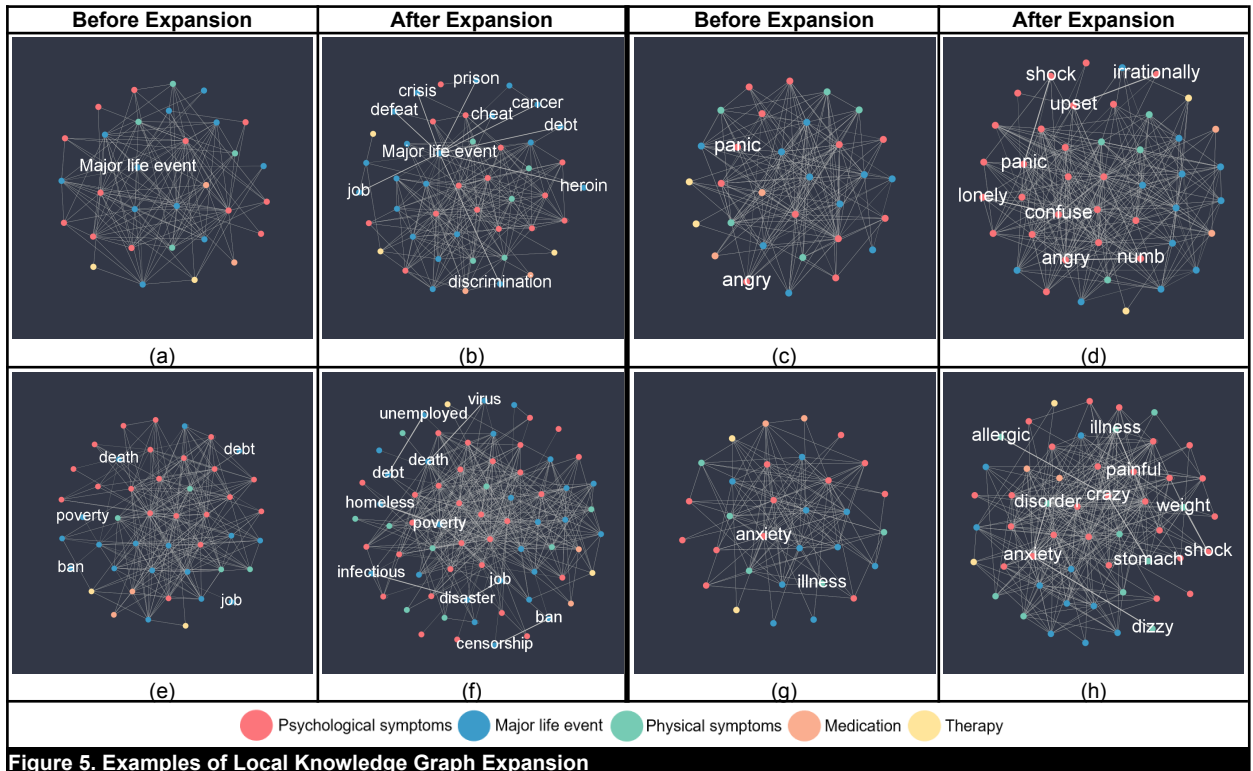

**Figure 5. Examples of Local Knowledge Graph Expansion**

We further assess whether knowledge derived from UGC provides meaningful contributions to the existing knowledge base by conducting a semi-structured interview with five mental health experts, including a chief physician, an associate chief physician, and three attending physicians from a national hospital. All interviewees have been working in the department of psychological medicine for at least 3 years and had a minimum of 5 years of experience providing healthcare services to depression patients. During the interviews, we presented our newly identified depression-related entities and triplets, and invited the health professionals to assess the clinical relevance, novelty, contribution to medical knowledge, information usefulness, and potential for clinical application of the discovered knowledge. Specifically, we ask interviewees to fill in the survey consisting of 5 questions about the knowledge contribution (7-point Likert scale) and one open question about their clinical suggestions (Table 6). All average scores were above 6, with clinical relevance, novelty, and contribution to medical literature at 6, information usefulness scoring 6.4, and the potential for clinical application receiving the highest score of 6.6. The chief physician provides feedback that the new depression knowledge provides important clues about depression patients' symptoms that the patient may forget to mention in clinical consultations. Our framework can assist large-scale and real-time surveillance of public mental health. The results enable health organizations to identify the potential impact of societal events, such as pandemic and economic downturn, on individuals' mental health.

| Table 6. Summary of Survey Results | | | | | | | |
|---|---|---|---|---|---|---|---|
| Responders | Professional Title | Years in Practice | Q1 | Q2 | Q3 | Q4 | Q5 |
| 1 | Chief Physician | 11–15 | 5 | 7 | 6 | 7 | 7 |
| 2 | Attending Physician | 5–10 | 6 | 7 | 5 | 7 | 7 |
| 3 | Resident Physician | 3–5 | 6 | 6 | 7 | 7 | 6 |
| 4 | Resident Physician | 3–5 | 7 | 7 | 7 | 5 | 6 |
| 5 | Resident Physician | 3–5 | 6 | 3 | 5 | 6 | 7 |
| Average | — | — | 6 | 6 | 6 | 6.4 | 6.6 |

**Note:** Overall Survey Instructions: You are presented with a list of depression symptoms, life events, and their associated relationships derived from user-generated content. For each question below, please rate your agreement with the statement on a scale from 1 (Strongly Disagree) to 7 (Strongly Agree).
Q1: **Clinical Relevance**: Are they clinically relevant to depression?
Q2: **Novelty**: They are not commonly captured in existing clinical studies or scientific literature.
Q3: **Contribution to Medical Knowledge**: Incorporating them could enhance the current medical knowledge base on depression.
Q4: **Usefulness of User-Generated Content**: User-generated content is a valuable source for identifying new symptoms, life events, and their associated relationships.
Q5: **Potential for Clinical Application**: They could be useful for clinical assessment, diagnosis, or monitoring of depression in patients.

### 4.4 Evaluation of the Depression Detection Performance

To demonstrate the effectiveness of our framework in depression detection, we compare it with three groups of strong benchmark methods: machine learning and feature engineering models (Chau et al., 2020; Choudhury et al., 2013; Coppersmith et al., 2014; Preoţiuc-Pietro et al., 2015; Reece et al., 2017), state-of-the-art end-to-end deep learning models (Benton et al., 2017; Ghosh et al., 2023; Khan et al., 2021; Lin et al., 2020; H. Liu et al., 2024; Malviya et al., 2021; Philip Thekkekara et al., 2024; W. Zhang et al., 2024), and large language models with fine tuning, zero shot learning, and few shot learning.

| Table 7. Comparison with Traditional Machine Learning with Feature Engineering | | | | | |
|---|---|---|---|---|---|
| | Model | Method | F1 | Precision | Recall |
| Machine | Choudhury et al. (2013) | Behavioral attributes, PCA, and SVM | 0.602 ± 0.001 | 0.824 ± 0.001 | 0.584 ± 0.001 |
| Learning and Feature Engineering | Coppersmith et al. (2014) | LIWC, uni-gram, character 5-gram, behavioral attributes, Log Linear classifier | 0.752 ± 0.001 | 0.788 ± 0.001 | 0.729 ± 0.001 |
| | Preotiuc-Pietro et al. (2015) | User attributes, LIWC, and LDA | 0.752 ± 0.001 | 0.803 ± 0.001 | 0.723 ± 0.001 |
| | Reece et al. (2017) | Linguistic feature and Random Forest | 0.739 ± 0.017 | 0.731 ± 0.020 | 0.748 ± 0.013 |
| | Chau et al. (2020) | Lexicon-based feature and Rule-based classification | 0.687 ± 0.005 | 0.672 ± 0.004 | 0.749 ± 0.005 |
| Deep Learning | Benton et al. (2017) | Multi-task learning | 0.753 ± 0.026 | 0.756 ± 0.024 | 0.752 ± 0.030 |
| | Lin et al. (2020) | CNN | 0.726 ± 0.006 | 0.721 ± 0.022 | 0.738 ± 0.015 |
| | Khan et al. (2021) | Bi-LSTM | 0.742 ± 0.010 | 0.759 ± 0.027 | 0.732 ± 0.017 |
| | Zhang et al. (2024) | LSTM & Entity recognition | 0.785 ± 0.013 | 0.823 ± 0.017 | 0.761 ± 0.024 |
| | Ghosh et al. (2023) | Attention-based Bi-LSTM and CNN | 0.732 ± 0.021 | 0.763 ± 0.048 | 0.718 ± 0.026 |
| | Thekkekara et al. (2024) | Attention-based CNN-BiLSTM | 0.764 ± 0.016 | 0.775 ± 0.021 | 0.756 ± 0.016 |
| | Liu et al. (2024) | Contrastive Learning | 0.668 ± 0.024 | 0.685 ± 0.047 | 0.669 ± 0.040 |
| | Malviya et al. (2021) | Transformer | 0.762 ± 0.011 | 0.783 ± 0.026 | 0.750 ± 0.021 |
| Large Language Models | GPT-4o-mini | Zero-shot learning * | 0.711 | 0.693 | 0.789 |
| | GPT-4o | | 0.778 | 0.827 | 0.747 |
| | GPT-4o-mini | Few-shot learning | 0.760 | 0.750 | 0.771 |
| | GPT-4o | | 0.772 | 0.801 | 0.751 |
| | DeepSeek-R1-Distill-Llama-8B | Fine-turning | 0.780 ± 0.012 | 0.801 ± 0.016 | 0.765 ± 0.017 |
| | Qwen3-8B | | 0.792 ± 0.008 | 0.812 ± 0.030 | 0.779 ± 0.022 |
| **Closed-loop Framework (Ours)** | | | **0.823 ± 0.005** | **0.853 ± 0.016** | **0.802 ± 0.016** |

Note: For a fair comparison, all benchmark methods were applied to the same dataset segment (eRisk 2018) over the same time period, with 60 percent allocated to training, 20 percent to validation, and 20 percent to testing.
Prompt for zero-shot and few-shot learning: *"Please analyze the following social media posts and determine whether the user is a depression patient. Output 1 if the user is a depression patient, and 0 if the user is not. <Example>"*

Our framework demonstrates significant improvements in predictive performance (Table 7). Compared with the best-performing traditional machine learning model (Preoțiuc-Pietro et al., 2015), it achieves a 7.1% increase in F1 score. Relative to the best-performing deep learning model (W. Zhang et al., 2024), it improves the F1 score by 3.8%. When compared with large language models, our framework demonstrates a clear advantage over zero-shot and few-shot learning, achieving a 4.5% improvement in F1 score relative to the state-of-the-art large language model GPT-4o under the zero-shot setting (our evaluation further indicates that including examples in the depression detection task using few-shot learning does not improve the performance of LLMs). In addition, when compared with the best-performing fine-tuned LLM, Qwen3-8B, our framework achieves a 3.1% improvement in F1 score.

Our proposed framework shows performance advantages in depression detection for several reasons. (1) Compared to traditional machine learning and feature engineering: the task is conducted at the user level rather than the post level, which is technically more complex, as it requires analyzing a large number of digital traces from UGC. However, extracting valid information from these traces poses a significant challenge for traditional machine learning models that depend on feature engineering. (2) Compared to deep learning methods, our proposed framework outperforms LSTM networks, attention-based models, contrastive learning models, and Transformer-based approaches in depression detection. The best-performing deep learning methods (W. Zhang et al., 2024) also integrate a static knowledge base and depression-related entities with a deep learning architecture. However, treating entity recognition and prediction as isolated tasks leads to suboptimal performance of each component and reduces the effectiveness of the overall model. This contrast highlights the advantages of the proposed closed-loop framework. (3) Compared to the direct use of LLMs, prevalent models such as GPT-4o and GPT-4o-mini with few-shot learning underperform relative to our proposed framework, indicating that pretrained LLMs alone cannot adequately address the depression detection task. Fine-tuned models, such as DeepSeek-R1-Distill-Llama-8 and Qwen3-8B, also lag behind our framework. Several factors may explain this gap. First, during fine-tuning on labeled UGC, LLMs simultaneously learn clinically relevant and irrelevant information. Without an intermediate, knowledge-guided task (for example, depression-related entity recognition), LLMs primarily capture associations between linguistic patterns and depression while neglecting underlying medical knowledge. Second, although LLMs can model the co-occurrence of words or concepts, they lack prior knowledge to leverage clinically meaningful relationships between depression-related entities and to quantify their importance for detection. For example, an LLM may recognize that "stomach pain" and "anxiety" frequently appear together in posts from depressed users, but it cannot infer that the co-occurrence of a mood symptom with a somatic symptom may support a depression diagnostic conclusion. In contrast, our closed-loop learning approach explicitly exploits these clinically meaningful relationships to generate more robust predictions. A further limitation of existing LLMs is their inability to systematically integrate useful medical knowledge extracted from UGC back into a knowledge base, either to enhance subsequent rounds of depression detection or to inform medical research. Collectively, these limitations highlight the advantages of our closed-loop framework. By combining LLMs with a knowledge graph, we not only leverage the strengths of pretrained LLMs but also systematically provide them with medical knowledge. At the same time, we use LLMs to expand the knowledge graph, thereby creating a mutually reinforcing closed-loop system.

## 4.5 Ablation and Sensitivity Analysis

We further conduct ablation and sensitivity analyses. For the ablation analysis, the closed-loop framework integrates depression-related entity recognition, depression knowledge graph learning, and depression detection within a joint learning design. To assess the contribution of each component, we successively remove the following: (1) the knowledge graph (the LLM module without entity embedding and entity importance), (2) knowledge graph refinement (the closed-loop system without knowledge graph refinement), (3) both the knowledge graph and depression-related entities (LLM alone), and (4) the joint learning design itself (a sequential model that first recognizes depression-related entities and then performs depression detection).

Table 8. Ablation Study

| Removed component | Method | F1 | Precision | Recall |
|---|---|---|---|---|
| Knowledge graph | LLM module | 0.805 ± 0.006 | 0.836 ± 0.025 | 0.784 ± 0.015 |
| Knowledge graph refinement | LLM module + Knowledge Graph module (without knowledge graph refinement) | 0.814 ± 0.008 | 0.835 ± 0.012 | 0.797 ± 0.008 |
| Knowledge graph + Depression-related entity recognition | LLM alone | 0.775 ± 0.005 | 0.781 ± 0.020 | 0.774 ± 0.017 |
| The joint learning design | Sequential model: depression-related entity recognition followed by depression detection | 0.791 ± 0.006 | 0.803 ± 0.010 | 0.781 ± 0.007 |
| Closed-loop Framework (Ours) | | **0.823 ± 0.005** | **0.853 ± 0.016** | **0.802 ± 0.016** |

Note: For a fair comparison, all models were applied to the same dataset segment (eRisk 2018) over the same time period, with 60 percent allocated to training, 20 percent to validation, and 20 percent to testing.

As shown in Table 8, removing any component, whether the knowledge graph, knowledge graph refinement, depression-related entity recognition, or the joint training architecture, significantly hampers detection performance. Specifically, removing the knowledge graph (LLM module without entity embedding and importance) and relying solely on recognized entities lowers the F1 score by 0.048. Incorporating the knowledge graph but excluding the refinement process reduces the F1 score by 0.036. Replacing the joint learning design with a sequential design for the depression-related entity recognition and depression detection tasks decreases the F1 score by 0.032. Eliminating both the knowledge graph and the depression-related entity recognition task results in a reduction of 0.018. Among different components, the knowledge graph is the most influential. The graph transition path-based entity importance (our technical contribution) enables the exploration of entity relations and the quantification of entity importance, which helps filter out irrelevant information from raw UGC. The refinement process further enhances embedding quality by adapting the graph knowledge to the context of UGC. Finally, the joint training design also improves overall model performance, providing additional support for our framework design.

In our closed-loop framework design, certain components involve alternative design choices. To validate that our selected configurations are optimal, we conduct three types of sensitivity analyses. The results are reported in Table 9. (1) Methods for estimating entity importance. To evaluate the effectiveness of our entity importance estimation design, we compare it against several alternative approaches, including Cosine Similarity (Carrasco & Rosillo, 2021), RGHAT path coefficients (Z. Zhang et al., 2020b), ConvE (Dettmers et al., 2018), and GraphSAGE (Hamilton et al., 2017). Our method consistently outperforms these path prediction approaches in terms of F1 score. (2) Strategies for joint learning. A major component of our method is joint learning, for which we evaluate several training strategies. First, we test loss function summation (Feng et al., 2022), where the overall loss is the sum of the depression-related entity recognition and depression detection losses. Second, we examine a task-specific approach (Zhao et al., 2018) that incorporates task-specific modules into a shared architecture. A projection layer connects the shared bottom layer with the task-specific output layers, allowing both shared and task-specific parameters. Finally, we test the mixture-of-experts model (Ma et al., 2018), where each task head receives a linear combination of shared feature outputs. Our current joint learning strategy, which alternates between training the depression detection task and the depression-related entity recognition task, outperforms the aforementioned alternative joint training strategies. (3) LoRa joint learning layer-splitting strategies. We further evaluate the sensitivity of the proposed framework to the tuning layers of the LLM using LoRA. The best F1 score (0.823) is achieved when layers 0-26 are tuned for the depression-related entity recognition task and layers 27-31 are tuned for the depression prediction task. This suggests that running more parameters in the depression-related entity recognition task helps capture the most informative entities for depression detection. Deeper layers (27-31) seem to be more depression detection task-specific, and shallow layers might perform more general content understanding. This underscores the necessity of partitioning tunable layers to optimize LLMs' capability for interrelated tasks.

Table 9. Sensitivity Analysis

| Category | Method | | F1 | Precision | Recall |
|---|---|---|---|---|---|
| Entity Importance Estimation | Cosine (Carrasco & Rosillo, 2021) | | 0.810 ± 0.011 | 0.838 ± 0.018 | 0.789 ± 0.010 |
| | ConvE matching score (Dettmers et al., 2018) | | 0.811 ± 0.012 | 0.830 ± 0.030 | 0.798 ± 0.012 |
| | RGHAT triplet weight (Zhang et al., 2020) | | 0.810 ± 0.011 | 0.841 ± 0.027 | 0.791 ± 0.024 |
| | GraphSAGE (Hamilton et al., 2017) | | 0.820 ± 0.012 | 0.848 ± 0.026 | **0.803 ± 0.029** |
| | **Ours: RGHAT triplet weight rescaled** | | **0.823 ± 0.005** | **0.853 ± 0.016** | 0.802 ± 0.016 |
| Joint Learning Strategy | Loss functions summation (Feng et al., 2022) | | 0.799 ± 0.004 | 0.840 ± 0.011 | 0.772 ± 0.005 |
| | Task-specific approach (Zhao et al., 2018) | | 0.804 ± 0.004 | **0.853 ± 0.008** | 0.771 ± 0.004 |
| | Mixture of experts (Ma et al., 2018) | | 0.800 ± 0.001 | 0.849 ± 0.007 | 0.768 ± 0.003 |
| | **Ours: Alternating training** | | **0.823 ± 0.005** | **0.853 ± 0.016** | **0.802 ± 0.016** |
| LoRa Joint Learning Layer Splitting Strategies | NER Layers: | Prediction Layers: | | | |
| | 0~24 | 25~31 | 0.809 ± 0.009 | 0.821 ± 0.028 | 0.801 ± 0.016 |
| | 0~28 | 29~31 | 0.813 ± 0.007 | 0.838 ± 0.021 | 0.795 ± 0.010 |
| | 0~4 | 5~31 | 0.800 ± 0.011 | 0.820 ± 0.009 | 0.785 ± 0.020 |
| | 0~31 | 0~31 | 0.806 ± 0.007 | 0.838 ± 0.024 | 0.783 ± 0.014 |
| | **Ours: 0~26** | **Ours: 27~31** | **0.823 ± 0.005** | **0.853 ± 0.016** | **0.802 ± 0.016** |

**Note:** NER Layers denote the LLM layers that are fine-tunable during the depression-related Entity Recognition task. Prediction Layers denote the LLM layers that are fine-tunable during the depression detection task.
For a fair comparison, all benchmark methods were applied to the same dataset segment (eRisk 2018) over the same time period, with 60 percent allocated to training, 20 percent to validation, and 20 percent to testing.

## 5 DISCUSSION

We present a closed-loop framework for simultaneous and continuous medical knowledge expansion and depression detection using UGC. Leveraging domain knowledge and LLMs, our framework accurately identifies individuals with depression based on UGC from social media and consistently outperforms strong baselines. Additionally, the framework can uncover new knowledge from emerging UGC, enhancing the existing knowledge base. This updated and enriched knowledge can be reintegrated into the framework, further improving prediction performance in subsequent prediction iterations. This enhanced domain knowledge also helps medical professionals understand depression-related factors and the characteristics of depression populations on social media, which are not easily identified in clinical settings.

From a design science perspective, we contribute an innovative deep learning framework that integrates knowledge graphs and LLMs through joint learning. Using depression detection on social media UGC as a case, the framework demonstrates adaptability to diverse contexts. It enhances prediction accuracy by embedding domain knowledge while simultaneously uncovering new insights. Beyond supporting social media platforms in identifying individuals at risk, the framework enriches existing knowledge of depression, e.g., offering medical professionals a broader understanding of depression-related factors often inaccessible in traditional clinical settings.

The proposed framework also holds significant implications for IS researchers. Domain knowledge is crucial for effective decision-making processes, and addressing issues related to knowledge discrepancy, discovery, and completion across various contexts is essential for enhancing these processes. By developing new machine learning techniques to tackle these challenges, researchers can help organizations and individuals fully utilize the computational power and data-driven insights of the AI and big data era. This, in turn, can lead to better-informed decisions and more efficient, effective actions. Additionally, knowledge discrepancy and completion are not only technical concerns but also carry ethical and societal implications. Ensuring that knowledge is comprehensive and accurate in diverse contexts is vital for responsible AI and data-driven decision-making.

## 6 CONCLUSION

Researchers have long recognized that integrating domain knowledge into machine learning can enhance the efficiency of disease detection using UGC. UGC can provide new insights related to certain diseases, which can be incorporated into existing medical domain knowledge to further enhance machine learning model performance on disease detection. These insights can contribute to a better understanding of disease-related factors in this new context, aiding medical professionals. In this paper, we emphasize the importance of incorporating domain knowledge to enhance machine learning model performance and propose an innovative framework that simultaneously identifies and uncovers new knowledge, thereby completing existing medical domain knowledge. We hope this research will inspire future researchers to focus on using machine learning methods to explore and improve existing domain knowledge.

D., Ramchandran, K., Zaharia, M., Gonzalez, J. E., & Stoica, I. (2025). *Why Do Multi-Agent LLM Systems Fail?* (No. arXiv:2503.13657). arXiv. https://doi.org/10.48550/arXiv.2503.13657

Chau, M., Li, T., Wong, P., Xu, J., Yip, P., & Chen, H. (2020). Finding People with Emotional Distress in Online Social Media: A Design Combining Machine Learning and Rule-Based Classification. *Management Information Systems Quarterly*, *44*(2), 933–955.

Chen, S., Liu, X., Gao, J., Jiao, J., Zhang, R., & Ji, Y. (2021). HittER: Hierarchical Transformers for Knowledge Graph Embeddings. In M.-F. Moens, X. Huang, L. Specia, & S. W. Yih (Eds.), *Proceedings of the 2021 Conference on Empirical Methods in Natural Language Processing* (pp. 10395–10407). Association for Computational Linguistics. https://doi.org/10.18653/v1/2021.emnlp-main.812

Chen, W., Li, W., Liu, X., Yang, S., & Gao, Y. (2023). Learning Explicit Credit Assignment for Cooperative Multi-Agent Reinforcement Learning via Polarization Policy Gradient. *Proceedings of the AAAI Conference on Artificial Intelligence*, *37*(10), 11542–11550. https://doi.org/10.1609/aaai.v37i10.26364

Choudhury, M. D., Gamon, M., Counts, S., & Horvitz, E. (2013, June 28). Predicting Depression via Social Media. *Seventh International AAAI Conference on Weblogs and Social Media*. Seventh International AAAI Conference on Weblogs and Social Media. https://www.aaai.org/ocs/index.php/ICWSM/ICWSM13/paper/view/6124

Coppersmith, G., Dredze, M., & Harman, C. (2014). Quantifying Mental Health Signals in Twitter. *Proceedings of the Workshop on Computational Linguistics and Clinical Psychology: From Linguistic Signal to Clinical Reality*, 51–60. https://doi.org/10.3115/v1/W14-3207

Cryan, J. F., & Dinan, T. G. (2012). Mind-altering microorganisms: The impact of the gut microbiota on brain and behaviour. *Nature Reviews Neuroscience*, *13*(10), 701–712. https://doi.org/10.1038/nrn3346

DeepSeek-AI, Guo, D., Yang, D., Zhang, H., Song, J., Zhang, R., Xu, R., Zhu, Q., Ma, S., Wang, P., Bi, X., Zhang, X., Yu, X., Wu, Y., Wu, Z. F., Gou, Z., Shao, Z., Li, Z., Gao, Z., … Zhang, Z. (2025). *DeepSeek-R1: Incentivizing Reasoning Capability in LLMs via Reinforcement Learning* (No. arXiv:2501.12948). arXiv. https://doi.org/10.48550/arXiv.2501.12948

Delmas, M., Wysocka, M., & Freitas, A. (2024). Relation Extraction in Underexplored Biomedical Domains: